\documentclass[10pt,twocolumn,letterpaper]{article}

\usepackage[pagenumbers]{cvpr}

\usepackage{cuted} 

\usepackage{currfile} 

\usepackage{caption} 

\usepackage{array}
\newcommand{\AccDense}{83.7}

\newcommand{\AccFtGentle}{83.0}

\newcommand{\AdvDelDense}{0.18}

\newcommand{\AdvDelFtDense}{0.18}
\newcommand{\AdvDelWandaNinety}{0.01}

\newcommand{\AdvDelWandaSeventy}{0.14}

\newcommand{\AdvInsDense}{0.20}

\newcommand{\AdvInsDiffWandaNinetyKd}{\ensuremath{+}0.03}

\newcommand{\AdvInsDiffWandaSeventyKd}{\ensuremath{+}0.05}
\newcommand{\AdvInsFtDense}{0.32}

\newcommand{\BaseBestMag}{66}
\newcommand{\BaseBestWanda}{65}

\newcommand{\BaseBreakLogpMag}{23}
\newcommand{\BaseBreakLogpWanda}{16}
\newcommand{\BaseBreakMag}{42}

\newcommand{\BaseBreakNWanda}{100}
\newcommand{\BaseBreakWanda}{25}

\newcommand{\BaseFpPrunedMax}{88}
\newcommand{\BaseFpPrunedMin}{60}
\newcommand{\BaseInvalidFpMax}{88}
\newcommand{\BaseInvalidFpMin}{60}
\newcommand{\BaseOwnMag}{82}
\newcommand{\BaseOwnWanda}{95}

\newcommand{\BestDense}{9.2}
\newcommand{\BestDiffNinety}{\ensuremath{+}3.32}
\newcommand{\BestDiffSeventy}{\ensuremath{+}1.31}
\newcommand{\BestHiNinety}{\ensuremath{+}6.03}
\newcommand{\BestHiSeventy}{\ensuremath{+}3.26}
\newcommand{\BestLoNinety}{\ensuremath{+}0.50}
\newcommand{\BestLoSeventy}{\ensuremath{-}0.26}
\newcommand{\BestN}{20}

\newcommand{\BinFidWandaNinety}{90.4}

\newcommand{\BinFrozenDropDense}{22.6}
\newcommand{\BinFrozenDropWandaNinety}{11.3}
\newcommand{\BinFrozenDropWandaSeventy}{19.2}

\newcommand{\BinFullAccWandaNinety}{82.6}

\newcommand{\CaseAccDense}{92}
\newcommand{\CaseAccOwn}{92}
\newcommand{\CaseAccPruned}{48}
\newcommand{\CaseCircuitPct}{15.8}
\newcommand{\CaseClass}{American alligator}
\newcommand{\CaseKappaDense}{0.72}
\newcommand{\CaseKappaModel}{Wanda 90\% (CE)}
\newcommand{\CaseKappaTarget}{0.68}
\newcommand{\CaseTrainedPct}{16.7}
\newcommand{\CaseUndoOther}{28}
\newcommand{\CaseUndoOwn}{79}

\newcommand{\CircKeptMaxMagNinety}{11.3}
\newcommand{\CircKeptMaxWandaNinety}{19.0}
\newcommand{\CircKeptMinMagNinety}{10.4}
\newcommand{\CircKeptMinWandaNinety}{13.4}

\newcommand{\CircKeptSizeRhoWandaNinety}{\ensuremath{-}0.95}

\newcommand{\CollapseVOneDense}{4}

\newcommand{\CollapseVOneTotal}{9}

\newcommand{\CollapseVOneWandaSeventy}{1}
\newcommand{\CollapseVOneWandaSeventyKd}{4}
\newcommand{\ConfDense}{0.78}
\newcommand{\ConfFtDense}{0.89}

\newcommand{\CoreChanceMax}{0.2}

\newcommand{\CoreCorrDiffWandaNinety}{\ensuremath{+}0.018}
\newcommand{\CoreCorrDiffWandaNinetyKd}{\ensuremath{+}0.017}
\newcommand{\CoreCorrDiffWandaSeventy}{\ensuremath{+}0.007}
\newcommand{\CoreCorrDiffWandaSeventyKd}{\ensuremath{+}0.004}

\newcommand{\CoreHundredDense}{11.9}

\newcommand{\CoreHundredWandaNinety}{25.0}

\newcommand{\CurveClampCINinetyKd}{\ensuremath{-}2.39 to \ensuremath{+}1.96}
\newcommand{\CurveClampCISeventyKd}{\ensuremath{+}0.37 to \ensuremath{+}1.18}
\newcommand{\CurveClampNinetyKd}{\ensuremath{+}0.29}
\newcommand{\CurveClampSeventyKd}{\ensuremath{+}0.77}

\newcommand{\CurveN}{20}
\newcommand{\CurveNativeCINinetyKd}{\ensuremath{-}5.69 to \ensuremath{+}0.56}
\newcommand{\CurveNativeCISeventyKd}{\ensuremath{-}1.01 to \ensuremath{+}0.60}
\newcommand{\CurveNativeNinetyKd}{\ensuremath{-}2.32}
\newcommand{\CurveNativeSeventyKd}{\ensuremath{-}0.18}
\newcommand{\CurveNativeSmallerNinetyKd}{9}
\newcommand{\CurveNativeSmallerSeventyKd}{6}

\newcommand{\DamageCtlRhoMax}{\ensuremath{+}0.50}
\newcommand{\DamageCtlRhoMin}{\ensuremath{+}0.19}

\newcommand{\DbAdvDelDense}{0.14}

\newcommand{\DbAdvDelWandaNinety}{-0.01}
\newcommand{\DbAdvDelWandaSeventy}{0.10}

\newcommand{\DbDzNeuronDiscFt}{\ensuremath{-}1.6}

\newcommand{\DbDzNeuronDiscWandaSeventy}{\ensuremath{-}0.5}

\newcommand{\DbDzSaeDiscFt}{\ensuremath{+}0.5}

\newcommand{\DbDzSaeDiscWandaSeventy}{\ensuremath{-}0.4}

\newcommand{\DbFtCmpDiffFt}{\ensuremath{+}0.47}
\newcommand{\DbFtCmpDiffWanda}{\ensuremath{+}1.57}

\newcommand{\DbMagCmpDiffMag}{\ensuremath{+}3.75}
\newcommand{\DbMagCmpDiffWanda}{\ensuremath{+}1.46}

\newcommand{\DecompGe}{10}
\newcommand{\DecompN}{14}

\newcommand{\DeitbEdgeDense}{23.4}
\newcommand{\DeitbEdgeFoldMax}{8.7}
\newcommand{\DeitbEdgeFoldMin}{3.1}
\newcommand{\DeitbEdgeShareDiffMax}{\ensuremath{+}3.5}

\newcommand{\DeitbEdgeShareDiffSeventyKd}{\ensuremath{+}2.0}

\newcommand{\DeitbProtUndoOtherMaxWanda}{37}

\newcommand{\DeitbProtUndoOwnWanda}{80}

\newcommand{\DeitbSlackDense}{58}

\newcommand{\DeitbSlackNinetyKd}{7}
\newcommand{\DeitbSlackSeventyKd}{35}

\newcommand{\DiagJaccardRepeat}{0.65}

\newcommand{\DiagNinetyOrig}{\ensuremath{+}7.8}
\newcommand{\DiagNinetyRepeat}{\ensuremath{+}7.3}
\newcommand{\DiagNinetyShuf}{\ensuremath{+}2.1}
\newcommand{\DiagNinetyShufTwo}{\ensuremath{+}2.3}

\newcommand{\DiagSeventyShuf}{\ensuremath{+}1.2}
\newcommand{\DiagSeventyShufTwo}{\ensuremath{+}1.2}

\newcommand{\DropAgreeNinetyAll}{\ensuremath{+}0.81}
\newcommand{\DropRelFtVsDense}{0.54}

\newcommand{\DropRelWandaNinety}{0.70}
\newcommand{\DropRelWandaSeventy}{0.49}
\newcommand{\DropSdFtVsDense}{5.3}

\newcommand{\DropSdWandaSeventy}{5.7}

\newcommand{\DzDenseInsCalWandaNinetyKd}{\ensuremath{-}0.4}

\newcommand{\DzDenseInsCalWandaSeventyKd}{\ensuremath{-}0.1}

\newcommand{\DzNeuronDiscWandaSeventy}{\ensuremath{+}0.6}

\newcommand{\DzSaeDiscFt}{\ensuremath{-}0.5}

\newcommand{\DzSaeDiscWandaSeventy}{\ensuremath{-}0.6}

\newcommand{\EdgeDense}{22.8}

\newcommand{\EdgeFoldNinety}{9}
\newcommand{\EdgeFoldSeventy}{3}

\newcommand{\EdgeShareDiffMax}{\ensuremath{+}2.0}
\newcommand{\EdgeShareDiffMin}{\ensuremath{+}0.9}

\newcommand{\EdgeSurvDense}{22.8}
\newcommand{\EdgeSurvWandaNinety}{24.9}
\newcommand{\EdgeSurvWandaSeventy}{24.2}

\newcommand{\ExactCoverMinMag}{91}

\newcommand{\ExactCoverMinWanda}{94}

\newcommand{\ExactCtlMag}{6}
\newcommand{\ExactCtlMagDeitb}{8}

\newcommand{\ExactCtlWanda}{8}
\newcommand{\ExactCtlWandaDeitb}{13}
\newcommand{\ExactGainHiMag}{1.03}

\newcommand{\ExactGainHiWanda}{0.75}

\newcommand{\ExactGainLoMag}{0.86}

\newcommand{\ExactGainLoWanda}{0.62}

\newcommand{\ExactNWanda}{100}

\newcommand{\ExactOwnMag}{81}
\newcommand{\ExactOwnMagDeitb}{69}
\newcommand{\ExactOwnWanda}{94}
\newcommand{\ExactOwnWandaDeitb}{79}
\newcommand{\ExactWinsMag}{100}

\newcommand{\ExactWinsWanda}{100}

\newcommand{\FkEightN}{24}

\newcommand{\FtCmpCIWandaMinusFt}{\ensuremath{+}0.28 to \ensuremath{+}0.89}
\newcommand{\FtCmpDiffFt}{\ensuremath{+}0.24}

\newcommand{\FtCmpDiffWandaMinusFt}{\ensuremath{+}0.58}
\newcommand{\FtCmpTargetKappa}{0.74}

\newcommand{\GentleGe}{4}
\newcommand{\GentleN}{6}

\newcommand{\InsCalDense}{0.770}
\newcommand{\InsCalFtDense}{0.790}

\newcommand{\InsCalWandaNinety}{0.577}

\newcommand{\InsCalWandaSeventy}{0.766}

\newcommand{\LamDenseSizeEight}{7.5}

\newcommand{\LamDenseSizeTwo}{20.6}

\newcommand{\MagCmpDiffMagSeventy}{\ensuremath{+}2.12}
\newcommand{\MagCmpDiffWandaSeventy}{\ensuremath{+}0.82}

\newcommand{\MeanDropWandaNinety}{20.1}

\newcommand{\NClassGroups}{13}

\newcommand{\NcutGapDense}{0.27}
\newcommand{\NcutGapFtDense}{0.27}

\newcommand{\NcutGapTopoWandaSeventy}{0.33}

\newcommand{\NcutGapWandaNinety}{1.03}
\newcommand{\NcutGapWandaSeventy}{0.54}

\newcommand{\NeuronDiscDense}{0.853}
\newcommand{\NeuronDiscFtDense}{0.773}

\newcommand{\NeuronDiscWandaNinety}{0.774}

\newcommand{\NeuronDiscWandaSeventy}{0.791}

\newcommand{\NeuronEntDense}{3.23}
\newcommand{\NeuronEntFtDense}{3.40}

\newcommand{\NeuronEntWandaNinety}{3.80}

\newcommand{\NeuronEntWandaSeventy}{3.39}

\newcommand{\NeuronRawDense}{0.351}
\newcommand{\NeuronRawFtDense}{0.458}

\newcommand{\NeuronRawWandaSeventy}{0.406}

\newcommand{\OrderAgreeShufDense}{89.1}

\newcommand{\OrderAgreeSortedDense}{87.4}

\newcommand{\OrderLambdaShufDense}{15.6}

\newcommand{\OrderLambdaSortedDense}{11.9}

\newcommand{\OrderSortedCoreDense}{1.09}

\newcommand{\OrderSortedCoreNinetyKd}{3.54}

\newcommand{\OrderSortedDiffNinetyKd}{\ensuremath{+}6.1}

\newcommand{\OrigSizeDiffWandaNinetyKd}{\ensuremath{+}1.46}

\newcommand{\OrigSizeDiffWandaSeventyKd}{\ensuremath{+}0.70}

\newcommand{\ProbeEarlyEightFtDense}{0.884}

\newcommand{\ProbeEarlyEightWandaNinety}{0.857}
\newcommand{\ProbeEarlyEightWandaSeventy}{0.880}

\newcommand{\ProbeEarlyOneFtDense}{0.782}
\newcommand{\ProbeEarlyOneMagSeventy}{0.759}
\newcommand{\ProbeEarlyOneWandaNinety}{0.702}
\newcommand{\ProbeEarlyOneWandaSeventy}{0.766}

\newcommand{\ProbeLateEightFtDense}{0.963}

\newcommand{\ProbeLateEightWandaNinety}{0.948}

\newcommand{\ProtAccNoneWanda}{70}

\newcommand{\ProtAccOwnWanda}{84}
\newcommand{\ProtBeatsEachMinBoth}{98}

\newcommand{\ProtMatchedFilled}{91}

\newcommand{\ProtOwnFracMag}{74}
\newcommand{\ProtOwnFracWanda}{73}

\newcommand{\ProtUndoOtherMax}{42}

\newcommand{\ProtUndoOwnMag}{81}
\newcommand{\ProtUndoOwnWanda}{93}

\newcommand{\ProtUndoRandMax}{22}

\newcommand{\ProtWeightsMag}{4.1}
\newcommand{\ProtWeightsWanda}{3.6}

\newcommand{\RestOwnHiMag}{87}
\newcommand{\RestOwnHiWanda}{101}
\newcommand{\RestOwnLoMag}{76}
\newcommand{\RestOwnLoWanda}{86}

\newcommand{\RndHeldoutDense}{52.5}

\newcommand{\RsN}{20}
\newcommand{\RsSizeCINinetyKd}{\ensuremath{-}1.07 to \ensuremath{+}1.25}
\newcommand{\RsSizeCISeventyKd}{\ensuremath{+}0.52 to \ensuremath{+}1.76}
\newcommand{\RsSizeDense}{19.8}
\newcommand{\RsSizeDiffNinetyKd}{\ensuremath{+}0.18}
\newcommand{\RsSizeDiffSeventyKd}{\ensuremath{+}1.09}

\newcommand{\SCoreDense}{2.66}

\newcommand{\SCoreRandRatio}{14}
\newcommand{\SCoreRandRef}{0.19}

\newcommand{\SCoreSizeDense}{0.177}

\newcommand{\SCoreSizeLocoMax}{\ensuremath{+}0.021}
\newcommand{\SCoreSizeLocoMin}{\ensuremath{+}0.001}

\newcommand{\SCoreSizeRelWandaNinetyKd}{11}

\newcommand{\SCoreSizeWandaNinety}{0.197}

\newcommand{\SCoreSizeWandaSeventy}{0.184}

\newcommand{\SCoreWandaNinety}{3.11}

\newcommand{\SFaithAgreeMax}{90.2}
\newcommand{\SFaithAgreeMin}{89.4}
\newcommand{\SFaithAgreeRandMax}{67.9}

\newcommand{\SLevelDiffMax}{\ensuremath{+}0.25}
\newcommand{\SLevelDiffMin}{\ensuremath{-}0.01}

\newcommand{\SLevelNMin}{71}

\newcommand{\SRepKappaDropWandaNinety}{0.042}
\newcommand{\SRepKappaDropWandaNinetyKd}{0.033}

\newcommand{\SRepN}{50}

\newcommand{\SSizeAgWandaNinetyKd}{14.2}

\newcommand{\SSizeDense}{15.0}
\newcommand{\SSizeDiffMax}{\ensuremath{+}0.84}
\newcommand{\SSizeDiffMin}{\ensuremath{+}0.57}

\newcommand{\SSizeWandaNinety}{15.8}
\newcommand{\SSizeWandaNinetyKd}{15.9}
\newcommand{\SSizeWandaSeventy}{15.8}

\newcommand{\STargetKappa}{0.69}

\newcommand{\SaeBlockFveFive}{0.9996}

\newcommand{\SaeBlockFveThree}{0.9998}
\newcommand{\SaeBlockLiveEleven}{14.1}
\newcommand{\SaeBlockLiveFive}{0.15}
\newcommand{\SaeBlockLiveNine}{11.3}
\newcommand{\SaeBlockLiveSeven}{0.79}
\newcommand{\SaeBlockLiveThree}{0.07}

\newcommand{\SaeBlockLzeroFive}{31}

\newcommand{\SaeBlockLzeroThree}{16}

\newcommand{\SaeDeadMax}{86}
\newcommand{\SaeDeadMin}{68}

\newcommand{\SaeDiscDense}{0.687}
\newcommand{\SaeDiscDenseRetrain}{0.687}
\newcommand{\SaeDiscFtDense}{0.673}

\newcommand{\SaeDiscWandaSeventy}{0.661}

\newcommand{\SaeSweepDeadAll}{37}
\newcommand{\SaeSweepFveAll}{0.948}
\newcommand{\SaeSweepFveCls}{0.993}
\newcommand{\SaeSweepFveExpMax}{0.9926}
\newcommand{\SaeSweepFveExpMin}{0.9925}

\newcommand{\SaeSweepFveKTwoFiftySix}{0.9949}
\newcommand{\SaeSweepFvePatch}{0.947}

\newcommand{\SepDense}{0.023}
\newcommand{\SepDenseHi}{0.033}
\newcommand{\SepDenseLo}{0.013}
\newcommand{\SepDiffBound}{0.008}

\newcommand{\SharpShareMax}{81}
\newcommand{\SharpShareMin}{60}

\newcommand{\SlackDense}{76}

\newcommand{\SwapHeldoutDense}{92.9}
\newcommand{\SwapHeldoutWandaNinety}{90.5}
\newcommand{\SwapNodeDense}{19.0}

\newcommand{\TinyClusterNodesMax}{1.5}
\newcommand{\TinyClusterShareDense}{20}
\newcommand{\TinyClusterShareWandaNinety}{40}
\newcommand{\TinyClusterShareWandaSeventy}{41}

\newcommand{\TrimDDense}{\ensuremath{-}7.3}
\newcommand{\TrimDNinety}{\ensuremath{-}7.1}
\newcommand{\TrimDSeventy}{\ensuremath{-}7.0}
\newcommand{\TrimHiDense}{\ensuremath{-}6.5}

\newcommand{\TrimLoDense}{\ensuremath{-}8.3}

\newcommand{\TrimNDense}{18}
\newcommand{\TrimNNinety}{16}
\newcommand{\TrimNSeventy}{14}

\newcommand{\UnitsOnlyDense}{7.3}
\newcommand{\UnitsOnlyDiffMax}{\ensuremath{+}1.52}
\newcommand{\UnitsOnlyDiffMin}{\ensuremath{+}0.45}

\newcommand{\VulnSizeRhoMagNinety}{\ensuremath{+}0.42}

\newcommand{\VulnSizeRhoWandaNinety}{\ensuremath{+}0.39}

\newcommand{\XlCircAttrRho}{0.15}
\newcommand{\XlCircAttrRhoCorr}{0.18}
\newcommand{\XlCircCIBound}{0.35}
\newcommand{\XlCircInside}{8}

\newcommand{\XlMatchedMaxRho}{0.13}

\newcommand{\XlMaxRhoCircuit}{0.15}

\newcommand{\XlNCircPairs}{12}

\newcommand{\XlNInside}{15}

\newcommand{\XlNPairs}{24}
\newcommand{\XlRelCirc}{0.89}
\newcommand{\XlRelDel}{0.82}
\newcommand{\XlRelIns}{0.79}
\newcommand{\XlRelNRepeat}{6}

\AtEndPreamble{
    \crefname{apptable}{App.\ Tab.}{App.\ Tabs.}
    \Crefname{apptable}{Appendix Table}{Appendix Tables}
    \crefname{appfigure}{App.\ Fig.}{App.\ Figs.}
    \Crefname{appfigure}{Appendix Figure}{Appendix Figures}
    \crefname{appendix}{App.}{Apps.}
    \Crefname{appendix}{Appendix}{Appendices}
}

\usepackage{subcaption}

\definecolor{cvprblue}{rgb}{0.21,0.49,0.74}
\usepackage[pagebackref,breaklinks,colorlinks,allcolors=cvprblue]{hyperref}

\hypersetup{pdftitle={Sparse but Not Simpler: Pruning Makes a Vision Transformer's Weights More Modular, but Not Its Computation},pdfauthor={Siyu Zhang}}

\title{Sparse but Not Simpler: Pruning Makes a Vision Transformer's Weights More Modular, but Not Its Computation}

\author{Siyu Zhang\\University of Texas at Austin\\Austin, TX 78712, USA}

\begin{document}
\maketitle

\begin{abstract}
Sparse networks are often expected to be easier to interpret, and language models trained with sparse weights have much smaller circuits than dense ones. We test whether pruning brings this benefit to vision transformers with IMPACT, an evaluation that compares dense and pruned models on weights, neurons, sparse-autoencoder features, class circuits and attribution maps, with a dense model finetuned without pruning as a control. Pruning makes the weights more modular, but no other level becomes simpler than the dense model. Decisions are compared through class circuits, the small sets of nodes behind the decision for one class, extracted so that circuits of different models are comparable. Restoring the weights that pruning removed from a class's dense circuit undoes most of that class's damage, far more than restoring as many weights elsewhere, so the circuits hold what each class depends on. Pruned circuits hold far fewer weights, but only because pruning removed weights everywhere; counted in nodes, they are no smaller and no less shared. Sparser or more modular weights are therefore no evidence of simpler decisions. At the other levels, finetuning alone changes the measures about as much as pruning does, so comparing the interpretability of compressed models needs a finetuned dense control.
\end{abstract}


\section{Introduction}
\label{sec:intro}

\begin{figure*}[t]
\centering
\begin{subfigure}[b]{2.0in}\centering
\includegraphics[width=2.0in]{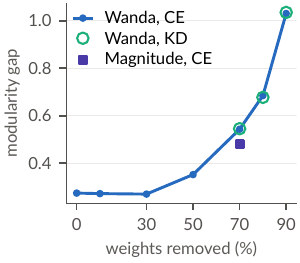}
\caption{Weights: more modular}
\label{fig:headline_a}
\end{subfigure}\hfill
\begin{subfigure}[b]{2.05in}\centering
\includegraphics[width=2.05in]{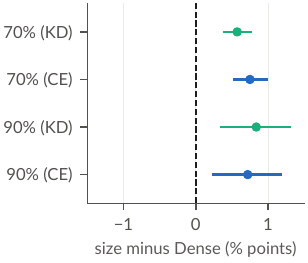}
\caption{Circuits: no smaller}
\label{fig:headline_b}
\end{subfigure}\hfill
\begin{subfigure}[b]{2.35in}\centering
\includegraphics[width=2.35in]{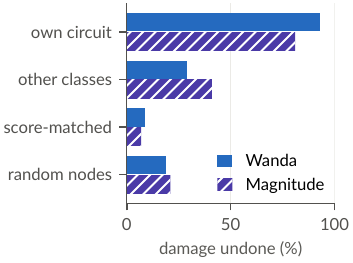}
\caption{Circuits: locate damage}
\label{fig:headline_c}
\end{subfigure}
\caption{\textbf{Pruning makes the weights more modular, but not the class circuits} (DeiT-III).
(a) Modularity gap of the MLP weight graphs (\cref{sec:res_weights}).
(b) Circuit size of recovered Wanda models minus Dense's, in percentage points of all nodes, at matched fidelity (\cref{sec:function}; 100 classes, 95\% intervals). CE: cross-entropy finetuning; KD: distillation.
(c) Share of the damage from one-shot 70\% pruning, summed over classes, undone by restoring equal numbers of weights from a class's own dense circuit, other classes' circuits or random nodes, or score-matched weights (at least $\ProtMatchedFilled\%$ as many; \cref{sec:damage}).}
\label{fig:headline}
\end{figure*}

Pruning can remove most of a trained vision transformer's weights with little loss in accuracy~\citep{sun2023simple}, which makes it an easy way to obtain sparse models. Sparse networks are often expected to be easier to interpret. Pruning splits the weight graph of a trained network into groups of neurons with strong connections within each group and weak connections between groups~\citep{filan2020pruned,filan2021clusterability}. Language models trained with sparse weights contain much smaller circuits, the sets of nodes that compute a decision, than dense models~\citep{gao2025weight}. If pruned vision transformers followed suit, every pruned model would have simpler decisions at no extra cost.

Interpretability is judged at several levels, from a model's weights to its input. The weights are judged by how modular their graph is~\citep{filan2021clusterability}, single neurons by how selectively they respond to a class~\citep{bau2017network}, and the features of sparse autoencoders (SAEs) in the same way, because SAEs separate concepts that single neurons mix~\citep{cunningham2023sparse}. Circuits are judged by how small and separate they are~\citep{olah2020zoom,bereska2024mechanistic}, and attribution maps by whether the pixels they mark drive the prediction~\citep{petsiuk2018rise}. A model can look simpler at one level and not at another. The class circuit of one class is a small set of nodes that still reproduces the model's decision about that class when every other node is replaced by its average activation. Class circuits test most directly whether decisions get simpler. We count a decision as simpler when its circuit has fewer nodes and shares fewer with other classes.

Earlier studies of pruned image classifiers examined individual units and found them no easier to interpret~\citep{frankle2019dissecting,vonrad2024pruning}. None of these studies examined whether pruning makes class circuits smaller or less shared. A circuit is worth measuring only if it captures what its class depends on. Pruning harms some classes far more than others~\citep{hooker2019forget,iofinova2023bias}, raising robustness and fairness concerns. If a circuit holds what a class depends on, the weights removed from it should explain the class's damage, so pruning tests the circuit.

Two obstacles complicate comparing the interpretability of dense and pruned models. First, pruned models are usually finetuned to recover accuracy, but finetuning changes a model even without pruning. Second, circuits are extracted by optimization, and the result depends on how the search is set up~\citep{drori2026unfaithful}. A circuit trained on the labels rather than the model's own decisions can follow rules the model does not use. Ablating nodes lets LayerNorm amplify the remaining ones. Circuits of different models also reproduce their models with different fidelity (\cref{app:extraction}). Apparent differences in circuit size therefore reflect the extraction as well as the models.

\begin{figure*}[t]
\centering
\includegraphics[width=6.5in]{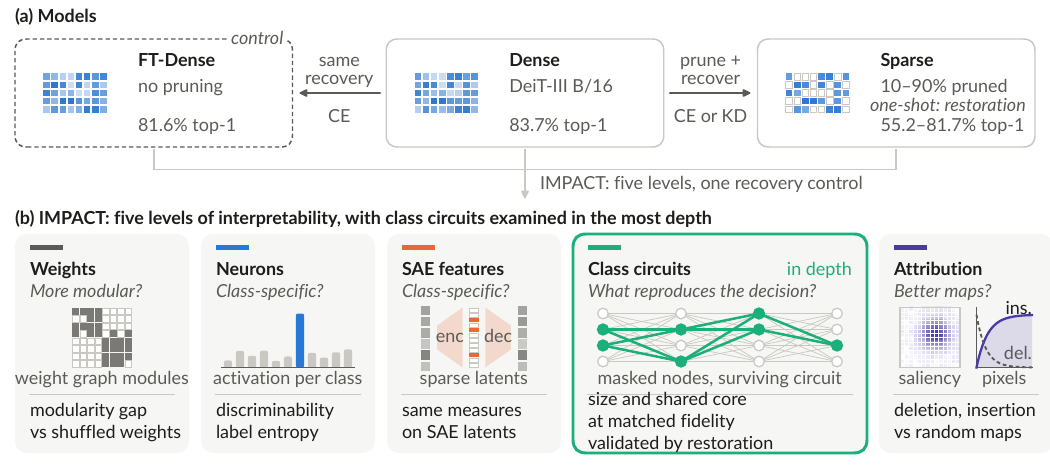}
\caption{\textbf{IMPACT compares models at five levels and examines class circuits in the most depth.}
(a) The dense model, its finetuned control (FT-Dense) and its sparse versions (accuracies: \cref{tab:recovery,tab:accuracy_sweep}).
(b) The five levels with their questions and measures.}
\label{fig:overview}
\end{figure*}

We build IMPACT (Interpretability Multi-level Pipeline for Assessing Computational Transparency), a multi-level evaluation that compares models on their weights, single neurons, SAE features, class circuits and attribution maps (\cref{fig:overview}). Every level is compared with a dense model finetuned without pruning (FT-Dense), which separates pruning from recovery, and every level except SAE features also with pruned models recovered through label-free distillation. For circuit comparisons, we reproduce each model's own decisions, limit LayerNorm amplification during node ablation, and compare circuits that reproduce their models equally well (matched fidelity; \cref{sec:circuits}). A restoration test then checks whether the circuits locate pruning damage. Our contributions on DeiT-III and DeiT-B are:

\begin{enumerate}
\item \textbf{IMPACT, a multi-level evaluation of interpretability across models} (\cref{sec:method,sec:overview}).

\item \textbf{Pruning makes weights more modular, but class circuits no smaller and no less shared} (\cref{sec:res_weights,sec:function}, \cref{fig:headline_a,fig:headline_b}). Pruned models' circuits contain far fewer weights only because the pruned models have fewer weights; they hold at least the same share of the surviving weights.

\item \textbf{Class circuits capture what each class depends on} (\cref{sec:damage}, \cref{fig:headline_c}). In a model pruned without recovery, putting back the weights removed inside a class's dense circuit reverses most of the aggregate drop in class log-probability, far more than putting back the same number of weights elsewhere. Classes with larger dense circuits also lose more accuracy.

\item \textbf{Recovery finetuning changes neuron, SAE and attribution measures as much as pruning does} (\cref{sec:recovery}). Pruning lowers deletion AUC, which would mean better attribution maps, but pruned models lose confidence when any pixels are deleted, even in random order. Changes in these measures show no clear relation with a class's circuit change.
\end{enumerate}

\section{Related work}
\label{sec:related}

\paragraph{Sparsity and interpretability.}
\citet{gao2025weight} trained small language models with most weights fixed at zero. On simple next-token tasks, the smallest circuit that reached a target loss had about $16\times$ fewer edges than in dense models of equal pretraining loss. \citet{yan2026swd} instead reparameterize pretrained weights into sparse factors and find smaller circuits at matched fidelity. Structured pruning removes whole tokens, heads or MLP dimensions of vision transformers~\citep{chavan2022vitslim,yang2023nvit}; sparse vision transformers have also been trained from scratch~\citep{chen2021chasing}, and Wanda has been applied to DeiT-B without retraining~\citep{sun2023simple}. We remove individual weights from a pretrained vision transformer, which preserves every node, and examine how its circuits change.

\paragraph{Structural and functional modularity.}
Structural modularity measures how cleanly the weight graph splits into clusters, using spectral clustering and shuffled-weight baselines~\citep{filan2020pruned,filan2021clusterability}. Pruning made multilayer perceptrons, but not small CNNs, more clusterable than their shuffles. Functional modularity instead concerns whether different functions use separate parts of a network and has been measured with weight masks learned per subtask~\citep{csordas2021modular}. The two can differ, as CNNs trained for clusterability had smaller class circuits but clusters that were no more specialized for classes~\citep{golechha2025modularity}, and in toy networks with built-in modules, specialization appeared only at extreme structural modularity~\citep{bena2025specialization}. Pruned subnetworks of a finetuned BERT also shared a core of attention heads~\citep{bhaskar2024heuristic}, which motivates our measurement of overlap across class circuits.

\paragraph{Circuit discovery.}
Circuits are found by activation patching or learned masks~\citep{conmy2023towards,bhaskar2024finding}, with benchmarks for language models~\citep{mueller2025mib}. Learned-mask circuits can use rules the model does not compute, exploit LayerNorm rescaling, or generalize differently from the model~\citep{drori2026unfaithful}. We address the first two (\cref{sec:circuits}). In vision transformers, class circuits were found by activation patching, with more overlap between related classes~\citep{zukowska2026seeing}. Concept circuits have been read from cross-layer transcoders~\citep{chen2026concept}, but neither study pruned models.

\paragraph{Disparate damage from pruning.}
Some classes, groups and atypical examples lose much more to pruning than others~\citep{hooker2019forget,hooker2020characterising,paganini2020prune}. Prior work traced this disparity to gradient norms and distance to the decision boundary~\citep{tran2022pruning} and to the dense model's uncertainty~\citep{iofinova2023bias}. These studies explain class-level damage through outputs and gradients.

\paragraph{Evaluation suites.}
Existing suites compare attribution methods, SAE architectures or circuit localization methods on one model at one level~\citep{hedstrom2023quantus,agarwal2022openxai,karvonen2025saebench,mueller2025mib}, and network dissection compares models by how well units align with labeled concepts~\citep{bau2017network}. Attribution maps for transformers come from relevance propagation~\citep{chefer2021transformer,ali2022xai,achtibat2024attnlrp} and are commonly scored by deletion and insertion~\citep{petsiuk2018rise,samek2016evaluating}. IMPACT compares models, controlling for recovery and circuit extraction.

\section{IMPACT}
\label{sec:method}

\subsection{Models and recovery control}
\label{sec:models}

To distinguish pruning from recovery finetuning, we compare pruned models with the original dense model (Dense) and with FT-Dense, which receives the same cross-entropy (CE) finetuning without pruning. We prune DeiT-III B/16~\citep{touvron2022deit}, trained on ImageNet-1k. Wanda~\citep{sun2023simple} removes, in each output row, the weights with the smallest product of magnitude and input activation; magnitude pruning~\citep{han2015deep} removes the smallest weights. We call the quantity by which each method ranks weights its pruning score. We prune to 70\% sparsity, where recovered top-1 stays within a few points of Dense, and to 90\%, where it falls by about twenty points (\cref{tab:recovery}; all sparsities: \cref{tab:accuracy_sweep}). Pruned models recover accuracy through CE finetuning on labels or knowledge distillation (KD) from the dense model without labels. We repeat the main comparisons on DeiT B/16 (DeiT-B)~\citep{touvron2021training}, using Wanda pruning to 70\% and 90\% sparsity followed by distillation or CE finetuning, with its own FT-Dense. Recipes are in \cref{app:impl}. Model names give method, sparsity and recovery, as in Wanda 70\% (CE).

\subsection{Weights}
\label{sec:weights}

Pruning's effect on the weights is measured as a \emph{modularity gap}, where larger means more modular. Following \citet{filan2021clusterability}, spectral clustering splits the graph of each MLP block (input, hidden and output units, with absolute weights as edges) into groups, and the cut sums the share of each group's edge weight that leaves the group. Because sparse graphs are easier to separate, the gap is the mean cut of copies whose weight matrices are each permuted entry by entry, zeros included (the same sparsity at random positions), minus the cut of the original graph. For pruned models, we also permute only the surviving weights among the surviving positions of each matrix, which keeps the pruning pattern (\cref{app:structure}).

\subsection{Comparable class circuits}
\label{sec:circuits}

A class circuit is a set of nodes that suffices to reproduce whether the full model predicts class $c$ for an image. A node is one of the model's activations at one of eight locations in every block: the LayerNorm outputs, the query, key and value of each attention head, the attention output, the MLP hidden units and the MLP output. The query, key and value have one node per head and the other locations one per dimension, 74{,}160 nodes in all (\cref{fig:nodes}). Every node outside the circuit is replaced by its mean activation at the same token position over the validation images (mean ablation). Every LayerNorm divides by at least the full model's scale (the LayerNorm clamp below). A circuit is thus evaluated inside its model rather than as a standalone subnetwork. To find a small set that preserves the decision, we follow \citet{gao2025weight} and learn a mask logit $m_i$ for each of the $N$ nodes, keeping node $i$ if $m_i>0$; a node's mask is shared across tokens and images. The logits minimize
\begin{equation}
\mathcal{L}(m)=\mathbb{E}_{x}\,\mathrm{CE}\big(z^{m}(x),\,y(x)\big)+\lambda\sum_{i=1}^{N}\sigma(m_i).
\label{eq:mask}
\end{equation}
Here $z^{m}(x)=[\max_{j\neq c}z^{m}_j(x),\,z^{m}_c(x)]$ pairs the logit of class $c$ with that of its strongest competing class in the masked model, so the circuit predicts $c$ exactly when $c$'s logit is the largest. The target $y(x)$ is one when the full model predicts $c$ and zero otherwise. The penalty, with $\sigma$ the sigmoid and $\lambda$ its weight, is a smooth count of retained nodes. The masks are binary (\cref{app:circuits}). We count circuit size in nodes, the units a reader must interpret.

Three choices make the extraction treat every model alike, each tested in \cref{app:extraction}. First, the target $y(x)$ is the model's own decision, not the label. Training against labels rewards a circuit for correcting the model's mistakes. Second, removing nodes shrinks the activations that LayerNorm normalizes, so LayerNorm amplifies the remaining ones and lets a few nodes compensate for those removed~\citep{drori2026unfaithful}. We limit this amplification with a \emph{LayerNorm clamp}. Every LayerNorm of the masked model divides by $\max(s_{\mathrm{clean}},s)$, where $s$ is its own scale and $s_{\mathrm{clean}}$ is the scale computed by the full model on the same input. Without the clamp, the dense and 70\%-sparse circuits lose far more fidelity than the 90\%-sparse ones (\cref{tab:circuits}, last column), so pruned circuits look smaller.

The third choice is to compare circuits at matched fidelity, because circuits trained with the same $\lambda$ reach different fidelity across models, and a circuit that gives up fidelity can be smaller. We measure \emph{fidelity} with Cohen's $\kappa$ between a circuit's class-vs-rest decisions and those of its model. Unlike raw agreement, $\kappa$ discounts agreement expected by chance, which matters when a model rarely predicts the class, as the 90\%-sparse models do. For each class, we first measure the fidelity of every compared model's trained circuit, and the lowest of these values becomes the common target fidelity (mean $\STargetKappa$). We then rank each model's nodes by mask logit and trim its circuit to the smallest size that reaches the target on the evaluation images (\cref{app:circuits}). For the \CaseClass{} class, Dense's trained circuit reaches $\kappa=\CaseKappaDense$ and the trained circuit with the lowest fidelity, that of \CaseKappaModel, reaches $\CaseKappaTarget$. Dense's circuit is therefore trimmed from $\CaseTrainedPct\%$ to $\CaseCircuitPct\%$ of the nodes. \cref{app:matched} repeats this selection on held-out halves. The trained circuits reproduce their models' decisions better than random circuits of equal size, agreeing with their models on $\SFaithAgreeMin$--$\SFaithAgreeMax\%$ of evaluation images, versus at most $\SFaithAgreeRandMax\%$ for random ones.

Sharing between class circuits is measured by the \emph{shared core}, the nodes present in at least 90\% of a model's class circuits. Larger circuits overlap more by chance, so we also divide core size by mean circuit size; a smaller ratio means more separation between class circuits.

We extract circuits for 100 ImageNet classes in every DeiT-III model compared. The classes were chosen by hand from $\NClassGroups$ semantic groups (\cref{app:classes}) before extraction. For DeiT-B, we use 20 of these classes (\cref{tab:classes}). We train every circuit on its class's training images and as many images of other classes, in one fixed shuffled order, and evaluate it on 50 validation images of the class and 50 of other classes. Circuit comparisons report paired differences from Dense with 95\% intervals from resampling classes; a difference is \emph{resolved} when its interval excludes zero.

\subsection{Validating circuits by restoration}
\label{sec:levels}

To test whether pruning damages a class through its circuit, we prune DeiT-III to 70\% (Wanda or magnitude) in one shot without recovery, so that the damage comes from the removed weights alone. We run a separate restoration experiment for each class, returning a subset of pruned weights to their dense values and measuring how much of the drop in mean class log-probability they reverse. A weight lies inside a set of nodes when both its input and output nodes belong to it (\cref{app:circuits}). Candidate weights come from the class's own dense circuit (as trained, before trimming to matched fidelity), from the circuits of three other classes, or from three random node sets with the own circuit's number of nodes per location. A fourth condition, the \emph{score-matched control}, draws weights from outside the circuit, matched to the own-circuit sample in layer and pruning score.

The own-circuit, other-class and random-node conditions restore equal numbers of weights, sampled uniformly. That number is the number of eligible pruned weights in the smallest of these sets, usually another class's circuit, in which case the own circuit is subsampled (\cref{app:circuits}). Restoring weights does not rebuild the circuit, and the rest of the model stays pruned.

\subsection{Neurons, features and attribution maps}
\label{sec:otherlevels}

Neurons are scored on the units that matter most for a class, in the last block (block 11), which is closest to the decision. For each class $c$, we select the 50 MLP units with the largest attribution-patching score $|\partial z_c/\partial h_u\cdot(h_u-\bar h_u)|$~\citep{nandaattribution}, which estimates how much replacing a unit's activation $h_u$ by its mean $\bar h_u$ changes the class logit $z_c$ (\cref{app:impl}). We ask how well each unit separates the class and how many classes it responds to. Let $A_u$ be the AUROC of the unit's CLS-token activation for class $c$ against other images, and $P_u(c')$ the share of the unit's peak absolute activation over patch tokens that falls on images of class $c'$. Then
\begin{equation}
\begin{aligned}
D(u)&=\max\{A_u,1-A_u\},\\
H(u)&=-\textstyle\sum_{c'}P_u(c')\log P_u(c'),
\end{aligned}
\label{eq:unit}
\end{equation}
A more interpretable unit has a higher \emph{discriminability} $D$, which runs from 0.5 (chance) to 1 (perfect separation), and a lower \emph{label entropy} $H$, which runs from 0 (one class) to $\log 1000\approx 6.91$ nats (all classes alike). The maximum scores units that suppress the class as well as those that promote it (\cref{app:suppressors}).

SAE features apply the same two measures to a learned sparse basis instead of the model's own units. For each model we train a BatchTopK SAE~\citep{cunningham2023sparse,bricken2023monosemanticity,bussmann2024batchtopk} on the CLS token after block 11 and score the 50 features with the largest mean activation on the class's images (\cref{app:sae}).

A good attribution map ranks first the pixels that drive the prediction. We compute maps with transformer attribution~\citep{chefer2021transformer} and, following \citet{petsiuk2018rise}, remove (deletion) or add (insertion) pixels in the map's order while recording the probability of the predicted class. The score is the area under this curve, where lower deletion and higher insertion mean a better map. Both depend on the model's confidence, so we divide insertion by the clean-image confidence $q$ and compare each map with a random pixel order on the same image (\cref{app:attribution}).

These measures have different units, so we report the effect size $d_z$, the mean per-class difference divided by its standard deviation (per image for attribution scores, over 2{,}000 evaluation images; \cref{app:stats}).

\section{Results}
\label{sec:overview}

\begin{table*}[t]
\centering
\caption{\textbf{Pruning makes the weights more modular, but no other level becomes simpler than the dense model} (DeiT-III; CE recovery). Arrows: the simpler direction. Circuit size: \% of nodes at matched fidelity (FT-Dense: $\Delta$ to Dense, matched with Dense and Wanda 70\% (CE), \cref{sec:function}). Weights: circuit weights as a share of the surviving weights. Deletion gain: random map's deletion AUC minus the map's. Cross-level: largest Spearman $|\rho|$ between per-class changes in circuit size and any other measure (Wanda 70\% (CE) vs FT-Dense; \cref{app:crosslevel}). --: not compared (\cref{app:coverage}).}
\label{tab:overview}
\footnotesize
\setlength{\tabcolsep}{4pt}
\begin{tabular}{lllcccc}
\toprule
Level & Question & Measure & Dense & FT-Dense & Wanda 70\% (CE) & Wanda 90\% (CE) \\
\midrule
Weights & More modular? & Modularity gap ($\uparrow$) & $\NcutGapDense$ & $\NcutGapFtDense$ & $\NcutGapWandaSeventy$ & $\NcutGapWandaNinety$ \\
\midrule
Neurons & More class-specific? & Discriminability ($\uparrow$) & $\NeuronDiscDense$ & $\NeuronDiscFtDense$ & $\NeuronDiscWandaSeventy$ & $\NeuronDiscWandaNinety$ \\
 & & Label entropy ($\downarrow$) & $\NeuronEntDense$ & $\NeuronEntFtDense$ & $\NeuronEntWandaSeventy$ & $\NeuronEntWandaNinety$ \\
SAE features & More class-specific? & Discriminability ($\uparrow$) & $\SaeDiscDense$ & $\SaeDiscFtDense$ & $\SaeDiscWandaSeventy$ & -- \\
\midrule
Class circuits & Smaller? & Size, \% of nodes ($\downarrow$) & $\SSizeDense$ & $\Delta$\,$\FtCmpDiffFt$ & $\SSizeWandaSeventy$ & $\SSizeWandaNinety$ \\
 & & Weights, \% of surviving ($\downarrow$) & $\EdgeSurvDense$ & -- & $\EdgeSurvWandaSeventy$ & $\EdgeSurvWandaNinety$ \\
 & Less shared? & Core/size ($\downarrow$) & $\SCoreSizeDense$ & -- & $\SCoreSizeWandaSeventy$ & $\SCoreSizeWandaNinety$ \\
 & Valid? & Restoration & \multicolumn{4}{>{\raggedright\arraybackslash}p{0.47\linewidth}}{one-shot 70\%, damage undone: own circuit $\ProtUndoOwnWanda\%$; other circuits, random nodes, score-matched weights $\le\ProtUndoOtherMax\%$; best class-specific score $\BaseBestWanda\%$ (own circuit, same budget: $\BaseOwnWanda\%$)} \\
\midrule
Attribution & Better maps? & Deletion gain over random ($\uparrow$) & $\AdvDelDense$ & $\AdvDelFtDense$ & $\AdvDelWandaSeventy$ & $\AdvDelWandaNinety$ \\
 & & Insertion$/q$ ($\uparrow$) & $\InsCalDense$ & $\InsCalFtDense$ & $\InsCalWandaSeventy$ & $\InsCalWandaNinety$ \\
\midrule
Across levels & Do changes agree? & Largest $|\rho|$ & \multicolumn{4}{l}{$\XlMaxRhoCircuit$, no correlation resolved} \\
\bottomrule
\end{tabular}
\end{table*}

Relative to the dense model, only the weights become simpler on every measure (\cref{tab:overview}). Units and SAE features become less discriminative, although SAE features respond to slightly fewer classes. For units, finetuning alone causes this loss and pruning recovers part of it (\cref{fig:decomposition}).

\subsection{Weights become more modular}
\label{sec:res_weights}

Pruning raises the modularity gap (\cref{fig:headline_a}), with Wanda doubling it at 70\% sparsity, from $\NcutGapDense$ to $\NcutGapWandaSeventy$, and nearly quadrupling it to $\NcutGapWandaNinety$ at 90\%. Magnitude pruning moves the same way (\cref{app:structure}). The dense weights are already more modular than shuffled weights, as \citet{filan2021clusterability} found for ImageNet CNNs, and finetuning does not change this. Against shuffles that keep the pruning pattern, the gap of Wanda 70\% is $\NcutGapTopoWandaSeventy$ instead of $\NcutGapWandaSeventy$, so most of the added structure lies in the pattern of removed weights.


\subsection{Circuits are no smaller}
\label{sec:function}

\begin{table*}[t]
\centering
\caption{\textbf{Pruned class circuits are slightly larger at matched fidelity} (DeiT-III, 100 classes). Size and shared core in \% of the nodes; $\Delta$: difference to Dense in points, $x^{+a}_{-b}$: 95\% interval from $x-b$ to $x+a$; classes smaller: number of classes with a smaller circuit than Dense's; $\kappa$ clamped$\to$unclamped: trained circuits' fidelity (before trimming) with and without the LayerNorm clamp.}
\label{tab:circuits}
\footnotesize
\setlength{\tabcolsep}{3pt}
\begin{tabular}{lcccccc}
\toprule
 & \multicolumn{5}{c}{Matched circuits} & Trained circuits \\
\cmidrule(lr){2-6}\cmidrule(lr){7-7}
Model & Size (\%) & $\Delta$ vs Dense & Classes smaller & Core (\%) & Core/size & $\kappa$ clamped$\to$unclamped \\
\midrule
Dense & 15.0 & -- & -- & 2.66 & 0.177 & 0.78$\to$0.63 \\
Wanda 70\% (KD) & 15.6 & $+0.57^{\mbox{\scriptsize$+0.21$}}_{\mbox{\scriptsize$-0.20$}}$ & 30/100 & 2.82 & 0.181 & 0.79$\to$0.64 \\
Wanda 70\% (CE) & 15.8 & $+0.75^{\mbox{\scriptsize$+0.25$}}_{\mbox{\scriptsize$-0.25$}}$ & 31/100 & 2.90 & 0.184 & 0.78$\to$0.66 \\
Wanda 90\% (KD) & 15.9 & $+0.84^{\mbox{\scriptsize$+0.47$}}_{\mbox{\scriptsize$-0.48$}}$ & 35/100 & 3.11 & 0.196 & 0.74$\to$0.71 \\
Wanda 90\% (CE) & 15.8 & $+0.72^{\mbox{\scriptsize$+0.47$}}_{\mbox{\scriptsize$-0.49$}}$ & 32/100 & 3.11 & 0.197 & 0.74$\to$0.70 \\
\bottomrule
\end{tabular}

\end{table*}

Counted in the weights that feed circuit nodes (a looser rule than in \cref{sec:levels}), pruned circuits do look far simpler. They hold about $\EdgeFoldSeventy$ and $\EdgeFoldNinety$ times fewer weights than Dense's circuits at 70\% and 90\% sparsity, but they do not occupy a smaller share of the weights that survive pruning (\cref{app:matched}).

At matched fidelity under the LayerNorm clamp, the four pruned Wanda models have circuits $\SSizeDiffMin$ to $\SSizeDiffMax$ percentage points of nodes larger than Dense's, with every interval above zero (\cref{fig:headline_b}). A dense class circuit contains $\SSizeDense\%$ of the nodes on average (\cref{tab:circuits}).

The recovery control separates two sources of this growth (\cref{tab:replications}). We match Dense, FT-Dense and Wanda 70\% (CE) alone, at a mean target $\kappa$ of $\FtCmpTargetKappa$ (higher than the $\STargetKappa$ of \cref{tab:circuits} because the match includes no 90\%-sparse model). FT-Dense then uses $\FtCmpDiffFt$ points more nodes than Dense, and Wanda 70\% (CE) uses another $\FtCmpDiffWandaMinusFt$ points beyond FT-Dense, both with intervals above zero. Finetuning alone enlarges the circuits slightly, and pruning enlarges them further. The distilled models also have larger circuits, so the growth does not require finetuning on labels. Magnitude pruning, which uses no data, enlarges them most, by $\MagCmpDiffMagSeventy$ points, more than twice Wanda's difference under the same recovery.

No alternative setting we tested gives a resolved size decrease, among them other sparsity penalties, fixed fidelity levels, a count of attention heads and MLP units alone, and DeiT-B (\cref{fig:robustness}). In DeiT-B, Wanda 70\% (CE) circuits are resolvably larger than Dense's ($\DbFtCmpDiffWanda$ points), but FT-Dense circuits are not ($\DbFtCmpDiffFt$). Pruned models keep less accuracy when random nodes are mean-ablated (\cref{sec:recovery}), so mean ablation could enlarge their circuits. Replacing ablated nodes with their activations on other images instead of their means keeps the distilled 70\%-sparse model's circuits larger than Dense's but leaves no resolved difference for the distilled 90\%-sparse model ($\RsN$ classes, \cref{app:matched}).

Relative to mean circuit size, the shared core changes little at 70\% sparsity (\cref{tab:circuits}). At 90\% it is about $\SCoreSizeRelWandaNinetyKd\%$ larger than Dense's, with an interval above zero (\cref{tab:matched_main}). The dense model's core contains $\SCoreDense\%$ of the nodes, about $\SCoreRandRatio$ times what random circuits of the same sizes would share (\cref{app:matched}). In Dense, circuits of classes from the same semantic group overlap more than those of different groups, as in \citet{zukowska2026seeing}, and pruning leaves this difference unchanged (\cref{app:matched}).

\subsection{Extraction choices matter}
\label{sec:extraction}

Five extraction choices change the dense-versus-pruned comparison (\cref{app:extraction}). Presenting the images in class-sorted order, all images of the class first, makes the circuits of the 90\%-sparse distilled model $\OrderSortedDiffNinetyKd$ points larger than Dense's on 20 of the classes, versus $\OrigSizeDiffWandaNinetyKd$ with a shuffled order. Under the sorted order, Dense's circuits end smaller and less faithful. An objective over all 1000 classes, a label target, evaluation without the clamp and matching raw agreement instead of $\kappa$ are the other four (\cref{tab:protocol}).

\subsection{Circuits locate pruning damage}
\label{sec:damage}

\begin{figure*}[t]
\centering
\includegraphics[width=5.85in]{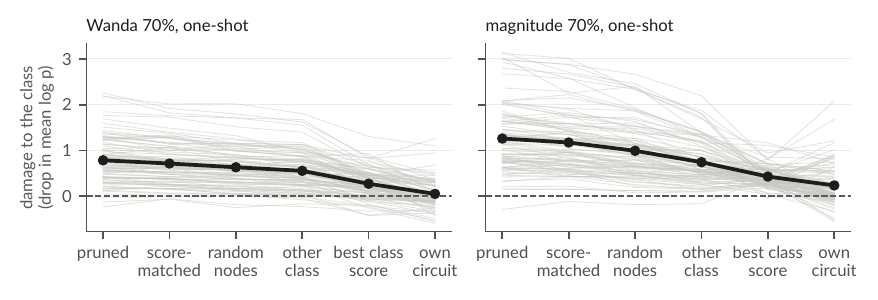}
\caption{\textbf{Restoring the weights in a class's own circuit undoes most of its damage} (100 classes, one-shot 70\% pruning). Damage: drop in the class's mean log-probability. Best class score: best of five class-specific scores (\cref{tab:restoration_baselines}). Each condition is a separate restoration; gray lines: one class; black: mean.}
\label{fig:protection}
\end{figure*}

The pruned weights restored inside a class's dense circuit (\cref{sec:levels}), a few percent of the blocks' linear weights, undo $\ProtUndoOwnWanda\%$ (Wanda) and $\ProtUndoOwnMag\%$ (magnitude) of the drop in class log-probability summed across classes (\cref{fig:protection}). The class with the largest drop is \CaseClass, whose trained dense circuit holds $\CaseTrainedPct\%$ of the nodes. Its accuracy falls from $\CaseAccDense\%$ to $\CaseAccPruned\%$ and returns to $\CaseAccOwn\%$ with its own circuit restored. Its own circuit undoes $\CaseUndoOwn\%$ of its log-probability drop, and other classes' circuits undo $\CaseUndoOther\%$.

Across classes, restoring the same number of weights from other classes' circuits or from random nodes undoes at most $\ProtUndoOtherMax\%$ of the damage, and so does the score-matched control. The own circuit beats each control in at least $\ProtBeatsEachMinBoth$ of the 100 classes. With exactly equal numbers of restored weights in every layer and pruning-score range, the own circuit still beats the score-matched control in every class for both methods. For Wanda it undoes $\ExactOwnWanda\%$ of the damage against $\ExactCtlWanda\%$ for the control (\cref{app:restoration}). In DeiT-B, the own circuit undoes $\DeitbProtUndoOwnWanda\%$ of Wanda's damage, against at most $\DeitbProtUndoOtherMaxWanda\%$ for other classes' circuits.

Weights chosen by class-specific importance scores come closer. These scores are computed on training images of the class: a Wanda score and four gradient scores, two from the dense and two from the pruned model. The best of the five, the Wanda score or a dense-model gradient score, undoes $\BaseBestWanda\%$ (Wanda) and $\BaseBestMag\%$ (magnitude), whereas a new sample of the own circuit at the same budget undoes $\BaseOwnWanda\%$ and $\BaseOwnMag\%$ (\cref{tab:restoration_baselines}). Unlike the own circuit, weights chosen by signed gradients on the pruned model make the model predict the class for other classes' images ($+\BaseFpPrunedMin$ to $+\BaseFpPrunedMax$ points), so they do not count as a repair (\cref{app:restoration}).

\subsection{Larger circuits, larger accuracy loss}
\label{sec:vulnerability}

\begin{figure}[t]
\centering
\includegraphics[width=\linewidth]{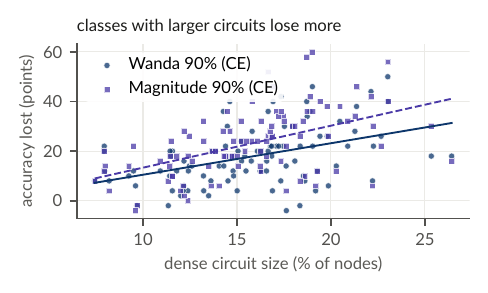}
\caption{\textbf{Classes with larger dense circuits lose more accuracy} under pruning to 90\% (Wanda and magnitude, CE recovery; 100 classes). Circuit size: nodes of the trained dense circuit; accuracy lost relative to FT-Dense.}
\label{fig:vulnerability}
\end{figure}

Beyond the restoration test, the size of a class's dense circuit is associated with its accuracy loss in recovered models pruned to 90\% (Spearman $\rho=\VulnSizeRhoWandaNinety$ for Wanda and $\VulnSizeRhoMagNinety$ for magnitude; \cref{fig:vulnerability}). For models with CE recovery, we measure this loss relative to FT-Dense, so that finetuning alone does not count as accuracy lost to pruning. At 90\% sparsity, Wanda with CE recovery costs a class $\MeanDropWandaNinety$ percentage points on average (reliability of the class differences: \cref{app:noise}). Wanda removes a larger share of a larger circuit's weights (Spearman $\rho=\CircKeptSizeRhoWandaNinety$ between circuit size and the share kept), whereas magnitude pruning keeps about the same share of every circuit (\cref{app:damage}). Both show the association (\cref{fig:vulnerability}), so it does not arise only because larger circuits lose more of their weights.

Circuit size remains associated with accuracy loss after controlling for six known factors, with partial Spearman correlations of $\DamageCtlRhoMin$ to $\DamageCtlRhoMax$ in the five pruned models (\cref{app:damage}). The factors are the dense model's gradient norm, distance to the decision boundary, uncertainty, number of training images, image entropy~\citep{paganini2020prune} and accuracy before pruning.

\subsection{Other levels move with recovery}
\label{sec:recovery}

\begin{table}[t]
\centering
\caption{\textbf{Recovery finetuning shifts predictions even without pruning} (DeiT-III; image sets: \cref{app:impl,app:drift}). KL: from the dense model's predictions; $\bar q$: mean confidence; Acc. kept: \% of top-1 kept with a random 10\% of nodes mean-ablated.}
\label{tab:recovery}
\footnotesize
\setlength{\tabcolsep}{2.5pt}
\begin{tabular}{lcccc}
\toprule
Model & Top-1 & KL & $\bar q$ & Acc. kept (\%) \\
\midrule
Dense & 83.7 & 0.00 & 0.78 & 76 \\
FT-Dense & 81.6 & 0.75 & 0.89 & 74 \\
\midrule
Wanda 70\% (KD) & 81.4 & 0.15 & 0.75 & 55 \\
Wanda 70\% (CE) & 80.4 & 0.72 & 0.86 & 50 \\
Magnitude 70\% (CE) & 80.3 & 0.72 & 0.86 & 50 \\
\midrule
Wanda 90\% (KD) & 62.6 & 1.03 & 0.52 & 1 \\
Wanda 90\% (CE) & 61.5 & 1.46 & 0.59 & 2 \\
\bottomrule
\end{tabular}

\end{table}

Although circuits show no simplification, neuron, SAE and attribution scores change when a model is pruned. We ask how much of that change comes from recovery finetuning instead. Recovery changes a model's predictions even without pruning, which shifts every measure that depends on them. Up to 70\% sparsity, CE recovery moves every model about the same distance from the dense model's predictions, whether or not it was pruned (KL in \cref{tab:recovery}). It does so mostly by raising confidence, from $\ConfDense$ to $\ConfFtDense$ (\cref{app:drift}). Distillation keeps the 70\%-sparse model close to the dense one, and at 90\% sparsity pruning moves the predictions further than recovery alone.

\begin{figure}[t]
\centering
\includegraphics[width=\linewidth]{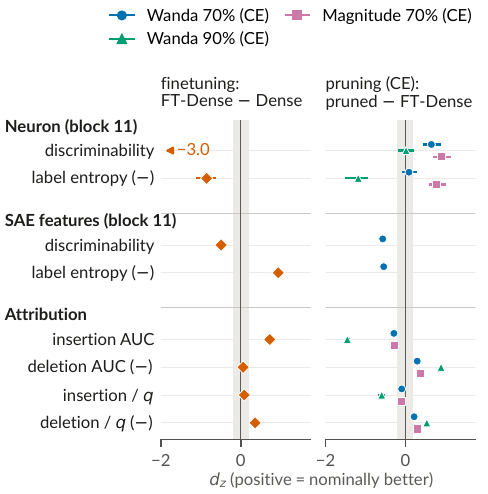}
\caption{\textbf{On most measures, finetuning alone has at least as large an effect as pruning to 70\%.} Effect sizes $d_z$ (95\% intervals) of finetuning (left, FT-Dense vs Dense) and of pruning (right, each model vs FT-Dense, CE recovery); finetuning's $|d_z|$ is at least pruning's in \DecompGe{} of \DecompN{} pairs at 70\% (SAE: Wanda only). Positive: the direction usually read as better; gray band: $|d_z|<0.2$.}
\label{fig:decomposition}
\end{figure}

On most neuron, SAE and attribution measures, finetuning alone (Dense to FT-Dense) has an effect size at least as large as that of pruning to 70\% under the same recovery recipe (FT-Dense to the pruned models; \cref{fig:decomposition}). Finetuning makes DeiT-III's block-11 units less discriminative and respond to more classes (\cref{tab:overview,tab:recovery_levels}). It also raises the insertion AUC of attribution maps, partly because the model becomes more confident.

Finetuning's effect also appears under a gentler recipe and, for units, in DeiT-B. A finetune that keeps top-1 within a point of Dense ($\AccFtGentle\%$ vs $\AccDense\%$) moves $\GentleGe$ of the $\GentleN$ measures at least as much as pruning does (\cref{tab:gentle}). On DeiT-B, finetuning again makes block-11 units less discriminative ($d_z=\DbDzNeuronDiscFt$). Pruning's effect on them reverses direction, from $\DzNeuronDiscWandaSeventy$ in DeiT-III to $\DbDzNeuronDiscWandaSeventy$ in DeiT-B, while its attribution effects keep their direction in both models. For SAE features the pattern is reversed. Pruning makes them less discriminative in both models ($d_z=\DzSaeDiscWandaSeventy$ and $\DbDzSaeDiscWandaSeventy$), whereas finetuning's effect reverses, from $\DzSaeDiscFt$ to $\DbDzSaeDiscFt$ (\cref{tab:deitb_recovery}).

Pruning lowers deletion AUC, which scores as better attribution maps, but it also lowers that of random maps (\cref{tab:attribution}). The deletion gain over random maps shrinks from $\AdvDelDense$ for Dense to $\AdvDelWandaNinety$ for Wanda 90\% (CE), and in DeiT-B from $\DbAdvDelDense$ to $\DbAdvDelWandaSeventy$ (70\%) and $\DbAdvDelWandaNinety$ (90\%; \cref{app:attribution,app:qualitative}). The lower deletion AUC therefore reflects fragility, not better explanations. Mean-ablating a random 10\% of nodes shows the same fragility, with 70\%-sparse models retaining about half their accuracy and 90\%-sparse models almost none, compared with $\SlackDense\%$ for Dense (\cref{tab:recovery}, \cref{fig:redundancy}).

If pruning simplified how a class is decided, the class's circuit and its other measures would change together. We compare Wanda 70\% (CE) with FT-Dense on the 100 classes so that the recovery recipe is fixed. For each class, we take the change in circuit size, measured as nodes and as weights of the trained circuit and as nodes at matched fidelity, and correlate it (Spearman) across classes with the change in each neuron, SAE and attribution measure. No correlation is resolved ($|\rho|\le\XlMaxRhoCircuit$; \cref{app:crosslevel}). Changes in the other measures show no clear relation to changes in a class's circuit.


\section{Discussion}
\label{sec:discussion}

Neither Wanda nor magnitude pruning scores weights by class, so neither rewards class-specific organization, which fits the lack of simpler circuits. By contrast, weight-sparse and clusterability training have produced smaller circuits~\citep{gao2025weight,golechha2025modularity}. We hypothesize that training can organize computation around sparsity or modularity, whereas pruning removes weights from a computation organized during dense training. Sparse factorization of pretrained language-model weights also yields smaller circuits~\citep{yan2026swd}. Pruning keeps the original units, whereas factorization creates new ones. Training the same architecture with sparse weights from scratch would test this hypothesis.

\paragraph{Limitations.}
Both models are supervised DeiT vision transformers trained on ImageNet, and both undergo unstructured pruning after training. Other architectures, self-supervised models, structured pruning and sparse training from scratch remain untested. We use distillation only with Wanda and test restoration only with dense circuits, after one-shot pruning to 70\%. Circuits cover 100 classes in DeiT-III and 20 in DeiT-B. Intervals resample classes, not recovery or extraction runs.

Circuit sizes depend on the extraction protocol. Absolute sizes change with the ablation, the sparsity penalty and trimming. Without the LayerNorm clamp, pruned circuits can look smaller than dense ones (\cref{sec:extraction}).

\section{Conclusion}
\label{sec:conclusion}

Pruning makes a vision transformer's weights more modular and its class circuits far smaller in weights, but only because pruning removed weights everywhere; counted in nodes, the circuits are no smaller. Post-training pruning is not a cheap route to explainable vision models, and neither weight counts nor weight modularity is evidence of simpler decisions. Dense circuits, found before pruning, identify the weights through which pruning harms each class. Recovery finetuning changes neurons, SAE features and attribution maps about as much as pruning does, and these changes show no clear relation to a class's circuit change. Comparing models by interpretability therefore requires controls for recovery, circuit extraction and fragility (\cref{sec:circuits,sec:recovery}).

\section*{LLM Usage Statement}
We used two large language models. Claude served in four roles: to write the code for the experiments, the analyses and the figures; to run and monitor the experiments; to give feedback on the drafts, including simulated reviews, consistency checks and checks of cited papers against their full text; and to revise the text. GPT reviewed the drafts and suggested revisions. The authors designed the study, decided which experiments to run and which claims to make, checked the code, the results and the text, and take full responsibility for the paper. The numbers, tables and data figures of the results are generated from the saved results by scripts, and a second script checks the paper's qualitative claims against those numbers.

{
    \small
    \bibliographystyle{ieeenat_fullname}
    \bibliography{main}
}

\clearpage
\onecolumn
\begin{center}
{\Large\textbf{Appendix}}
\end{center}
\appendix
\crefalias{section}{appendix}
\crefalias{table}{apptable}
\crefalias{figure}{appfigure}
\crefname{apptable}{Tab.}{Tabs.}
\crefname{appfigure}{Fig.}{Figs.}

Citations, and references to Sections 1--6, Figures 1--5 and Tables 1--3, refer to the main paper.
\crefname{appendix}{Appendix}{Appendices}

\section{Implementation details}
\label{app:impl}

\paragraph{Neuron metrics.}
For each class, we select the 50 block-11 MLP units with the largest attribution-patching score~\citep{nandaattribution}, $|\partial z_c/\partial h_u \cdot (h_u-\bar h_u)|$ for the class logit $z_c$ and the mean activation $\bar h_u$, on the class's validation images; a class's score is the mean of a measure over these 50 units. \cref{app:blocks} repeats the unit measures at blocks 2 and 7. SAE features are the 50 latents with the largest mean activation on the class's images. Discriminability and label entropy are computed on a balanced set of training images (built as in \cref{app:circuits}, Training order). Label entropy uses the maximum absolute activation of a unit over patch tokens, so that a unit responding anywhere in the image, with either sign, counts as active. It is measured in nats, at most 6.91 for 1000 classes. For a unit $u$ with this peak $a_u(x)$ on image $x$, label entropy is $H(u)=-\sum_c P_u(c)\log P_u(c)$ with $P_u(c)\propto\sum_{x\in D_c}a_u(x)$ for the images $D_c$ of class $c$.

\paragraph{Pruning and recovery.}
Wanda scores every weight of every linear layer, including the classifier, by $|W_{ij}|\,\|X_j\|_2$ over the calibration images and zeroes the lowest-scoring fraction of each output row; magnitude pruning ranks by $|W_{ij}|$ alone.

Wanda calibration uses one validation image per class for the CE and BCE models and one training image per class for the KD models; the two calibration sets give identical pruning decisions for 98.8\% of the weights of the 70\%-sparse model. The 1{,}000 validation calibration images are among the 50{,}000 evaluation images; Wanda uses only their activations, never their labels.

All recoveries train for two epochs on the ImageNet training set with AdamW (weight decay 0.01) at a constant learning rate of $10^{-5}$ (gentle CE: $10^{-6}$) and batch size 32, with random resized crops and horizontal flips, keeping the pruning masks fixed.

BCE applies binary cross-entropy to the one-hot labels, as in DeiT-III's own training. Distillation minimizes $\mathrm{KL}(p_{\mathrm{dense}}\,\|\,p_{\mathrm{pruned}})$ at temperature 1 and does not use the labels; distilling the dense model into itself has zero loss and would change it only through weight decay, so the dense model is its own distillation control.

\paragraph{Transformer attribution for DeiT-III.}
DeiT-III scales each residual branch with a learned LayerScale vector, $y=\gamma\odot h$. We add the relevance rule $R_h=R_y\odot y/(y+\epsilon\,\mathrm{sign}(y))$, which passes relevance through unchanged up to the stabilizer $\epsilon$ because the scaling is element-wise. We prepend a zero row to the positional embedding for the CLS token. Maps are computed for a fixed random subset of 2{,}000 validation images, because they are costly to compute.

\paragraph{Accuracy kept.} A random fraction of the nodes at every mask location is replaced by their mean, and top-1 accuracy on the 5{,}000 validation images of \cref{app:drift} (five per class) is recorded relative to the model's own, averaged over three random draws. \cref{tab:recovery} reports the fraction 10\%, where the dense and the 70\%-sparse models differ most; \cref{fig:redundancy} shows all fractions.

\section{Predictions, accuracy and robustness after recovery}
\label{app:drift}
\cref{tab:drift} lists, for every model, the KL divergence of its predictions from the dense model's, the share of images on which its top-1 prediction agrees with the dense model's, and its mean confidence, on five validation images per class. Dividing the logits by a fitted temperature removes $\SharpShareMin$--$\SharpShareMax\%$ of the KL in every label-finetuned model up to 70\% sparsity, with CE or with DeiT-III's own BCE loss ($T$ and KL ($T$ fit) in \cref{tab:drift}). Label finetuning raises confidence above the dense model's at every sparsity up to 70\%.

\begin{table}[h]
\centering
\caption{\textbf{How far recovery moves the predictions from the dense model's} (5{,}000 validation images). KL: $\mathrm{KL}(p_{\mathrm{dense}}\,\|\,p_{\mathrm{model}})$; Agree: top-1 agreement with Dense; $\bar q$: mean confidence; $T$: the temperature that, dividing the model's logits, minimizes the KL (searched from 0.25 to 4); KL ($T$ fit): the KL left at that temperature.}
\label{tab:drift}
\footnotesize
\begin{tabular}{lcccccc}
\toprule
Model & KL & KL ($T$ fit) & $T$ & Agree (\%) & $\bar q$ \\
\midrule
Dense & 0.00 & -- & -- & 100.0 & 0.78 \\
FT-Dense (gentle) & 0.59 & 0.12 & 1.55 & 93.6 & 0.90 \\
FT-Dense & 0.75 & 0.20 & 1.62 & 90.0 & 0.89 \\
FT-Dense (BCE) & 1.30 & 0.25 & 2.05 & 90.7 & 0.93 \\
Wanda 30\% (CE) & 0.72 & 0.19 & 1.62 & 90.6 & 0.89 \\
Wanda 50\% (CE) & 0.72 & 0.20 & 1.62 & 90.4 & 0.89 \\
Wanda 70\% (KD) & 0.15 & 0.15 & 1.00 & 90.7 & 0.75 \\
Wanda 70\% (CE) & 0.72 & 0.29 & 1.55 & 87.8 & 0.86 \\
Wanda 70\% (BCE) & 1.18 & 0.30 & 1.91 & 88.4 & 0.90 \\
Magnitude 70\% (CE) & 0.72 & 0.29 & 1.52 & 88.5 & 0.86 \\
Wanda 90\% (KD) & 1.03 & 1.03 & 1.02 & 67.0 & 0.52 \\
Wanda 90\% (CE) & 1.46 & 1.25 & 1.38 & 65.1 & 0.59 \\
\bottomrule
\end{tabular}

\end{table}

\begin{table}[h]
\centering
\caption{\textbf{Neuron and attribution measures of every model} (DeiT-III). Discr., Entr.: discriminability and label entropy of block-11 units; Ins./$q$: insertion AUC over confidence; Rand.\ del.: deletion AUC of random maps (lower = more fragile).}
\label{tab:recovery_levels}
\footnotesize
\begin{tabular}{lcccc}
\toprule
 & \multicolumn{2}{c}{Neurons} & \multicolumn{2}{c}{Attribution} \\
\cmidrule(lr){2-3}\cmidrule(lr){4-5}
Model & Discr. & Entr. & Ins./$q$ & Rand. del. \\
\midrule
Dense & 0.853 & 3.23 & 0.770 & 0.394 \\
FT-Dense (gentle) & 0.798 & 3.29 & 0.821 & 0.436 \\
FT-Dense & 0.773 & 3.40 & 0.790 & 0.385 \\
\midrule
Wanda 70\% (KD) & 0.818 & 3.55 & 0.738 & 0.301 \\
Wanda 70\% (CE) & 0.791 & 3.39 & 0.766 & 0.317 \\
Wanda 70\% (BCE) & 0.804 & 3.34 & 0.769 & 0.340 \\
Magnitude 70\% (CE) & 0.796 & 3.25 & 0.764 & 0.299 \\
\midrule
Wanda 90\% (KD) & 0.739 & 3.86 & 0.599 & 0.082 \\
Wanda 90\% (CE) & 0.774 & 3.80 & 0.577 & 0.093 \\
\bottomrule
\end{tabular}

\end{table}

\begin{table}[h]
\centering
\caption{\textbf{Finetuning lowers unit discriminability in both models; pruning's effect on it reverses.} Effect sizes $d_z$ (95\% intervals) of finetuning (FT-Dense vs Dense) and of pruning (Wanda 70\% (CE) vs FT-Dense); neurons and SAE features: 100 classes, attribution: 2{,}000 images. Positive: the direction usually read as better; ($-$): sign reversed.}
\label{tab:deitb_recovery}
\footnotesize
\setlength{\tabcolsep}{2.5pt}
\begin{tabular}{@{}llcccc@{}}
\toprule
 & & \multicolumn{2}{c}{DeiT-III} & \multicolumn{2}{c}{DeiT-B} \\
\cmidrule(lr){3-4}\cmidrule(l){5-6}
Level & Measure & Finetuning & Pruning & Finetuning & Pruning \\
\midrule
Neuron (block 11) & discriminability & $-3.01^{\mbox{\scriptsize$+0.40$}}_{\mbox{\scriptsize$-0.56$}}$ & $+0.65^{\mbox{\scriptsize$+0.25$}}_{\mbox{\scriptsize$-0.22$}}$ & $-1.60^{\mbox{\scriptsize$+0.29$}}_{\mbox{\scriptsize$-0.38$}}$ & $-0.55^{\mbox{\scriptsize$+0.19$}}_{\mbox{\scriptsize$-0.19$}}$ \\
Neuron (block 11) & label entropy ($-$) & $-0.85^{\mbox{\scriptsize$+0.22$}}_{\mbox{\scriptsize$-0.28$}}$ & $+0.09^{\mbox{\scriptsize$+0.19$}}_{\mbox{\scriptsize$-0.19$}}$ & $+1.09^{\mbox{\scriptsize$+0.30$}}_{\mbox{\scriptsize$-0.24$}}$ & $+0.27^{\mbox{\scriptsize$+0.21$}}_{\mbox{\scriptsize$-0.18$}}$ \\
Neuron (block 11) & class variance ($-$) & $-0.34^{\mbox{\scriptsize$+0.19$}}_{\mbox{\scriptsize$-0.20$}}$ & $+0.54^{\mbox{\scriptsize$+0.23$}}_{\mbox{\scriptsize$-0.20$}}$ & $-0.08^{\mbox{\scriptsize$+0.18$}}_{\mbox{\scriptsize$-0.21$}}$ & $-0.20^{\mbox{\scriptsize$+0.20$}}_{\mbox{\scriptsize$-0.20$}}$ \\
SAE features & discriminability & $-0.49^{\mbox{\scriptsize$+0.06$}}_{\mbox{\scriptsize$-0.06$}}$ & $-0.57^{\mbox{\scriptsize$+0.06$}}_{\mbox{\scriptsize$-0.07$}}$ & $+0.47^{\mbox{\scriptsize$+0.22$}}_{\mbox{\scriptsize$-0.20$}}$ & $-0.41^{\mbox{\scriptsize$+0.18$}}_{\mbox{\scriptsize$-0.20$}}$ \\
SAE features & label entropy ($-$) & $+0.94^{\mbox{\scriptsize$+0.07$}}_{\mbox{\scriptsize$-0.07$}}$ & $-0.54^{\mbox{\scriptsize$+0.06$}}_{\mbox{\scriptsize$-0.07$}}$ & $+0.72^{\mbox{\scriptsize$+0.23$}}_{\mbox{\scriptsize$-0.18$}}$ & $+0.02^{\mbox{\scriptsize$+0.19$}}_{\mbox{\scriptsize$-0.19$}}$ \\
SAE features & class variance ($-$) & $+0.08^{\mbox{\scriptsize$+0.05$}}_{\mbox{\scriptsize$-0.06$}}$ & $-0.68^{\mbox{\scriptsize$+0.07$}}_{\mbox{\scriptsize$-0.08$}}$ & $-0.44^{\mbox{\scriptsize$+0.19$}}_{\mbox{\scriptsize$-0.20$}}$ & $+0.51^{\mbox{\scriptsize$+0.22$}}_{\mbox{\scriptsize$-0.19$}}$ \\
Attribution & insertion AUC & $+0.73^{\mbox{\scriptsize$+0.06$}}_{\mbox{\scriptsize$-0.06$}}$ & $-0.29^{\mbox{\scriptsize$+0.05$}}_{\mbox{\scriptsize$-0.05$}}$ & $+0.58^{\mbox{\scriptsize$+0.06$}}_{\mbox{\scriptsize$-0.05$}}$ & $-0.48^{\mbox{\scriptsize$+0.04$}}_{\mbox{\scriptsize$-0.05$}}$ \\
Attribution & deletion AUC ($-$) & $+0.06^{\mbox{\scriptsize$+0.04$}}_{\mbox{\scriptsize$-0.05$}}$ & $+0.29^{\mbox{\scriptsize$+0.05$}}_{\mbox{\scriptsize$-0.04$}}$ & $-0.09^{\mbox{\scriptsize$+0.04$}}_{\mbox{\scriptsize$-0.05$}}$ & $+0.42^{\mbox{\scriptsize$+0.05$}}_{\mbox{\scriptsize$-0.05$}}$ \\
Attribution & insertion / $q$ & $+0.09^{\mbox{\scriptsize$+0.05$}}_{\mbox{\scriptsize$-0.04$}}$ & $-0.09^{\mbox{\scriptsize$+0.04$}}_{\mbox{\scriptsize$-0.05$}}$ & $+0.00^{\mbox{\scriptsize$+0.04$}}_{\mbox{\scriptsize$-0.05$}}$ & $-0.19^{\mbox{\scriptsize$+0.05$}}_{\mbox{\scriptsize$-0.07$}}$ \\
Attribution & deletion / $q$ ($-$) & $+0.36^{\mbox{\scriptsize$+0.05$}}_{\mbox{\scriptsize$-0.04$}}$ & $+0.22^{\mbox{\scriptsize$+0.05$}}_{\mbox{\scriptsize$-0.05$}}$ & $+0.24^{\mbox{\scriptsize$+0.05$}}_{\mbox{\scriptsize$-0.04$}}$ & $+0.29^{\mbox{\scriptsize$+0.07$}}_{\mbox{\scriptsize$-0.07$}}$ \\
\bottomrule
\end{tabular}

\end{table}

\begin{table}[h]
\centering
\caption{\textbf{A gentler finetune moves most measures at least as much as pruning} (DeiT-III). Effect sizes $d_z$ (95\% intervals) of the gentle finetune (learning rate $10^{-6}$, top-1 $\AccFtGentle\%$ vs $\AccDense\%$ for Dense) and of the standard finetune, both vs Dense, and of pruning (Wanda 70\% (CE) vs FT-Dense). Positive: the direction usually read as better; ($-$): sign reversed.}
\label{tab:gentle}
\footnotesize
\setlength{\tabcolsep}{2.5pt}
\begin{tabular}{@{}llccc@{}}
\toprule
Level & Measure & Gentle finetune & Finetune & Pruning \\
\midrule
Neuron (block 11) & discriminability & $-2.29^{\mbox{\scriptsize$+0.35$}}_{\mbox{\scriptsize$-0.54$}}$ & $-3.01^{\mbox{\scriptsize$+0.40$}}_{\mbox{\scriptsize$-0.56$}}$ & $+0.65^{\mbox{\scriptsize$+0.25$}}_{\mbox{\scriptsize$-0.22$}}$ \\
Neuron (block 11) & label entropy ($-$) & $-0.43^{\mbox{\scriptsize$+0.20$}}_{\mbox{\scriptsize$-0.20$}}$ & $-0.85^{\mbox{\scriptsize$+0.22$}}_{\mbox{\scriptsize$-0.28$}}$ & $+0.09^{\mbox{\scriptsize$+0.19$}}_{\mbox{\scriptsize$-0.19$}}$ \\
Attribution & insertion AUC & $+1.12^{\mbox{\scriptsize$+0.09$}}_{\mbox{\scriptsize$-0.08$}}$ & $+0.73^{\mbox{\scriptsize$+0.06$}}_{\mbox{\scriptsize$-0.06$}}$ & $-0.29^{\mbox{\scriptsize$+0.05$}}_{\mbox{\scriptsize$-0.05$}}$ \\
Attribution & deletion AUC ($-$) & $-0.16^{\mbox{\scriptsize$+0.05$}}_{\mbox{\scriptsize$-0.04$}}$ & $+0.06^{\mbox{\scriptsize$+0.04$}}_{\mbox{\scriptsize$-0.05$}}$ & $+0.29^{\mbox{\scriptsize$+0.05$}}_{\mbox{\scriptsize$-0.04$}}$ \\
Attribution & insertion / $q$ & $+0.25^{\mbox{\scriptsize$+0.05$}}_{\mbox{\scriptsize$-0.06$}}$ & $+0.09^{\mbox{\scriptsize$+0.05$}}_{\mbox{\scriptsize$-0.04$}}$ & $-0.09^{\mbox{\scriptsize$+0.04$}}_{\mbox{\scriptsize$-0.05$}}$ \\
Attribution & deletion / $q$ ($-$) & $+0.19^{\mbox{\scriptsize$+0.05$}}_{\mbox{\scriptsize$-0.05$}}$ & $+0.36^{\mbox{\scriptsize$+0.05$}}_{\mbox{\scriptsize$-0.04$}}$ & $+0.22^{\mbox{\scriptsize$+0.05$}}_{\mbox{\scriptsize$-0.05$}}$ \\
\bottomrule
\end{tabular}

\end{table}

\cref{tab:accuracy_sweep} gives the top-1 accuracy of every CE-recovered Wanda and magnitude model on the 50{,}000 validation images, including Magnitude 90\% (CE), which enters only the class-damage analysis.

\begin{table}[h]
\centering
\caption{\textbf{Top-1 accuracy after CE recovery} (\%, DeiT-III, 50{,}000 validation images), by share of weights removed.}
\label{tab:accuracy_sweep}
\footnotesize
\begin{tabular}{lccccccc}
\toprule
Pruning & 0\% & 10\% & 30\% & 50\% & 70\% & 80\% & 90\% \\
\midrule
Wanda (CE) & 83.7 & 81.5 & 81.7 & 81.6 & 80.4 & 76.9 & 61.5 \\
Magnitude (CE) & 83.7 & 81.5 & 81.7 & 81.6 & 80.3 & 76.6 & 55.2 \\
FT-Dense (CE) & 81.6 &  &  &  &  &  &  \\
\bottomrule
\end{tabular}

\end{table}

Distillation and 90\% sparsity show effects beyond finetuning. In distilled models, the label entropy of units rises with sparsity. Under CE recovery, by contrast, the 70\%-sparse model's entropy is no higher than FT-Dense's, so its increase over Dense comes from finetuning alone. At 90\% sparsity, confidence-normalized insertion AUC falls with either recovery (\cref{tab:recovery_levels}).

\begin{figure}[h]
\centering
\includegraphics[width=3.12in]{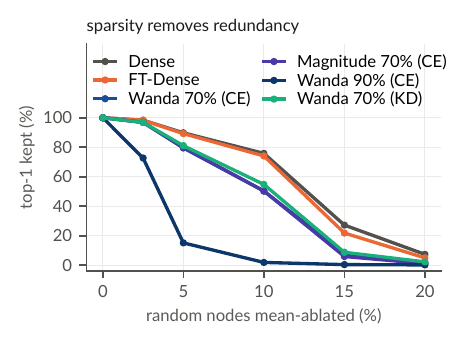}
\caption{\textbf{Pruned models keep less accuracy when nodes are ablated.} Accuracy kept with a random fraction of nodes replaced by their means (Wanda and magnitude 70\% coincide).}
\label{fig:redundancy}
\end{figure}

Accuracy kept under ablation (\cref{app:impl}) follows the sparsity whatever the recovery (\cref{sec:recovery}), and it also falls with sparsity in DeiT-B, from $\DeitbSlackDense\%$ to $\DeitbSlackSeventyKd\%$ at 70\% and $\DeitbSlackNinetyKd\%$ at 90\% (distillation; \cref{fig:redundancy}).

\section{Matched circuits}
\label{app:matched}
\cref{tab:matched_main} gives every paired difference to Dense under our protocol on the 100 classes, with 95\% intervals from resampling classes (the shared core is recomputed in every resample) and, in parentheses, the $p$-value of a paired permutation test that swaps the two models' circuits within randomly chosen classes (10{,}000 permutations, $p=(\text{extreme}+1)/(\text{draws}+1)$, uncorrected).

In the finetuning control of \cref{tab:replications}, Wanda 70\% (CE) needs $\FtCmpDiffWandaMinusFt$ points more than FT-Dense (\FtCmpCIWandaMinusFt). In DeiT-B, magnitude pruning to 70\% (CE) also enlarges the circuits most, by $\DbMagCmpDiffMag$ points against $\DbMagCmpDiffWanda$ for Wanda 70\% (CE) (20 classes, matched among the three models).

Matching raw agreement instead of $\kappa$ gives wider intervals at 90\% sparsity, where chance agreement is high (\cref{tab:matched_checks}).

At fixed $\kappa$ levels between 0.4 and 0.6 (reached by at least $\SLevelNMin$ of the 100 classes in every model), the pruned models' cores stay within $\SLevelDiffMin$ to $\SLevelDiffMax$ points of Dense's. At thresholds of 80\% and 95\%, every pruned model's core also exceeds Dense's.

Because the core is a threshold statistic and the bootstrap duplicates classes, we also leave out one class at a time: the core/size differences then lie between $\SCoreSizeLocoMin$ and $\SCoreSizeLocoMax$.

The held-out columns select each model's circuit on one random half of every class's validation images and evaluate it on the other half, for $\SRepN$ stratified splits in both directions; classes are the resampling unit. In the held-out check, the 90\%-sparse models' circuits are larger but also slightly less faithful than the dense model's ($\kappa$ lower by $\SRepKappaDropWandaNinetyKd$ and $\SRepKappaDropWandaNinety$).

Random circuits that keep each class circuit's number of nodes per location but place them at random share $\SCoreRandRef\%$ of the nodes in at least 90\% of the classes, compared with $\SCoreDense\%$ for the dense model's circuits. Computed the same way for every model, this chance core holds at most $\CoreChanceMax\%$ of the nodes. Subtracting it from each model's core leaves the core/size differences from Dense at $\CoreCorrDiffWandaSeventy$ and $\CoreCorrDiffWandaSeventyKd$ at 70\% sparsity (CE, KD) and $\CoreCorrDiffWandaNinety$ and $\CoreCorrDiffWandaNinetyKd$ at 90\%.

\begin{figure}[h]
\centering
\includegraphics[width=3.25in]{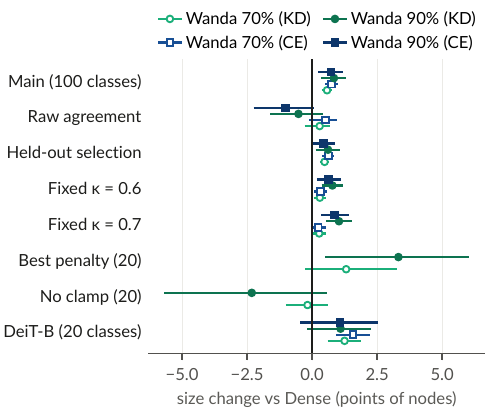}
\caption{\textbf{No check shows a resolved reduction.} Wanda circuit size minus Dense's (points of nodes; 95\% intervals). Fixed $\kappa$: classes reaching the level in every model. Best penalty: smallest circuit over three penalties (distilled models, 20 classes). Raw agreement: matched on agreement instead of $\kappa$. Held-out selection: chosen on half of each class's images, scored on the other half. No clamp: without the LayerNorm clamp (distilled models, 20 classes). DeiT-B Wanda 70\% (CE): matched with Dense and FT-Dense (\cref{tab:replications}).}
\label{fig:robustness}
\end{figure}

No check gives a resolved size decrease (\cref{fig:robustness}). Negative size estimates appear only with raw-agreement matching at 90\% sparsity and without the LayerNorm clamp, and their intervals include zero. In DeiT-B (20 classes), pruned circuits are larger on average, with two of the four intervals excluding zero.

\cref{tab:separation} asks whether circuits of related classes separate after pruning and where the extra nodes sit. For each model we compute the Jaccard similarity of the matched circuits of every pair of classes; the separation $S$ is the mean over pairs from the same semantic group minus the mean over pairs from different groups, with the same paired class resampling as above. In Dense, $S=\SepDense$ ($\SepDenseLo$ to $\SepDenseHi$), so the measure sees semantic structure; no pruned model changes it by more than $\SepDiffBound$ in either direction of its interval. The extra nodes lie mostly in MLP units, attention outputs and, at 70\% sparsity, the LayerNorm outputs; at 90\% the LayerNorm outputs lose nodes and the MLP units supply nearly all of the increase. Sparse probes nonetheless find no spreading of class information over more units up to 70\% sparsity (\cref{app:probe}).

Counting only attention heads and MLP units, which unlike residual-stream dimensions are units of the model's own basis, Dense circuits hold $\UnitsOnlyDense\%$ of them and the pruned models' circuits $\UnitsOnlyDiffMin$ to $\UnitsOnlyDiffMax$ points more, every interval above zero.

Pruned models also lose more accuracy when random nodes are mean-ablated (\cref{sec:recovery}). Models that keep equal accuracy under ablation still differ in circuit size (Magnitude and Wanda 70\% (CE), $\MagCmpDiffMagSeventy$ and $\MagCmpDiffWandaSeventy$ points). The 90\%-sparse models, which keep almost no accuracy, have circuits about as enlarged as the 70\%-sparse ones under the main protocol (\cref{tab:circuits}) but more enlarged at fixed fidelity and at fixed penalty (\cref{tab:fixedkappa,tab:lambda}), as fragility would predict. Under resample ablation, where removed nodes take the full model's activation on an image of another class, Dense circuits on the $\RsN$ original classes hold $\RsSizeDense\%$ of the nodes; Wanda 70\% (KD) needs $\RsSizeDiffSeventyKd$ points more (\RsSizeCISeventyKd) and Wanda 90\% (KD) $\RsSizeDiffNinetyKd$ (\RsSizeCINinetyKd), against $\OrigSizeDiffWandaSeventyKd$ and $\OrigSizeDiffWandaNinetyKd$ under mean ablation. The extra enlargement at 90\% sparsity thus comes with mean ablation, whereas the enlargement at 70\% persists.

A weight counts toward a circuit when the node it feeds is in the circuit (for the MLP output weights, when both of their nodes are); the restoration test uses the stricter rule that both nodes of every weight lie in the node set (\cref{app:circuits}). Counted in weights, the pruned models' circuits hold $\EdgeShareDiffMin$ to $\EdgeShareDiffMax$ points more of their surviving weights than Dense's circuits hold of the dense model's ($\EdgeDense\%$), every difference resolved but small. In DeiT-B (20 classes), the pruned models' circuits have $\DeitbEdgeFoldMin$ to $\DeitbEdgeFoldMax$ times fewer weights than Dense's ($\DeitbEdgeDense\%$ of the dense model's weights); their share of the surviving weights is $\DeitbEdgeShareDiffSeventyKd$ points higher at 70\% and up to $\DeitbEdgeShareDiffMax$ points higher at 90\%.

\begin{table*}[t]
\centering
\caption{\textbf{Separation by semantic group and extra nodes by location} (100 classes, matched $\kappa$). $S$: mean Jaccard similarity of same-group minus different-group class circuits; $\Delta S$: change from Dense, $x^{+a}_{-b}$: 95\% interval from $x-b$ to $x+a$; extra nodes: mean number of nodes per class circuit minus Dense's, at each mask location.}
\label{tab:separation}
\footnotesize
\begin{tabular}{lcccccccccc}
\toprule
 & \multicolumn{2}{c}{Separation $S$} & \multicolumn{8}{c}{Extra nodes vs Dense, per location} \\
\cmidrule(lr){2-3}\cmidrule(lr){4-11}
Model & $S$ & $\Delta S$ & LN1 & Q & K & V & Attn out & LN2 & MLP units & MLP out \\
\midrule
Dense & 0.023 & -- & -- & -- & -- & -- & -- & -- & -- & -- \\
Wanda 70\% (KD) & 0.024 & $+0.001^{\mbox{\scriptsize$+0.004$}}_{\mbox{\scriptsize$-0.003$}}$ & \ensuremath{+}42 & \ensuremath{+}3 & \ensuremath{-}1 & \ensuremath{+}6 & \ensuremath{+}57 & \ensuremath{+}84 & \ensuremath{+}160 & \ensuremath{+}74 \\
Wanda 70\% (CE) & 0.023 & $+0.000^{\mbox{\scriptsize$+0.004$}}_{\mbox{\scriptsize$-0.003$}}$ & \ensuremath{+}76 & \ensuremath{+}5 & \ensuremath{+}0 & \ensuremath{+}7 & \ensuremath{+}121 & \ensuremath{+}84 & \ensuremath{+}205 & \ensuremath{+}59 \\
Wanda 90\% (KD) & 0.022 & $+0.000^{\mbox{\scriptsize$+0.006$}}_{\mbox{\scriptsize$-0.007$}}$ & \ensuremath{-}49 & \ensuremath{-}3 & \ensuremath{+}4 & \ensuremath{+}9 & \ensuremath{+}141 & \ensuremath{-}97 & \ensuremath{+}555 & \ensuremath{+}61 \\
Wanda 90\% (CE) & 0.024 & $+0.001^{\mbox{\scriptsize$+0.007$}}_{\mbox{\scriptsize$-0.006$}}$ & \ensuremath{-}62 & \ensuremath{-}1 & \ensuremath{+}5 & \ensuremath{+}9 & \ensuremath{+}130 & \ensuremath{-}123 & \ensuremath{+}536 & \ensuremath{+}39 \\
\bottomrule
\end{tabular}

\end{table*}

\begin{table}[h]
\centering
\caption{\textbf{Main protocol: differences to Dense at matched $\kappa$} (100 classes). Size and core in points of the nodes, core/size as a ratio; $x^{+a}_{-b}$: 95\% interval from $x-b$ to $x+a$; parentheses: permutation $p$.}
\label{tab:matched_main}
\footnotesize
\begin{tabular}{lccc}
\toprule
Model & Size & Core & Core/size \\
\midrule
Wanda 70\% (KD) & $+0.57^{\mbox{\scriptsize$+0.21$}}_{\mbox{\scriptsize$-0.20$}}$ (\ensuremath{<}0.001) & $+0.16^{\mbox{\scriptsize$+0.07$}}_{\mbox{\scriptsize$-0.11$}}$ (0.001) & $+0.004^{\mbox{\scriptsize$+0.003$}}_{\mbox{\scriptsize$-0.007$}}$ \\
Wanda 70\% (CE) & $+0.75^{\mbox{\scriptsize$+0.25$}}_{\mbox{\scriptsize$-0.25$}}$ (\ensuremath{<}0.001) & $+0.24^{\mbox{\scriptsize$+0.06$}}_{\mbox{\scriptsize$-0.15$}}$ (\ensuremath{<}0.001) & $+0.007^{\mbox{\scriptsize$+0.002$}}_{\mbox{\scriptsize$-0.009$}}$ \\
Wanda 90\% (KD) & $+0.84^{\mbox{\scriptsize$+0.47$}}_{\mbox{\scriptsize$-0.48$}}$ (0.001) & $+0.45^{\mbox{\scriptsize$+0.16$}}_{\mbox{\scriptsize$-0.21$}}$ (\ensuremath{<}0.001) & $+0.019^{\mbox{\scriptsize$+0.007$}}_{\mbox{\scriptsize$-0.012$}}$ \\
Wanda 90\% (CE) & $+0.72^{\mbox{\scriptsize$+0.47$}}_{\mbox{\scriptsize$-0.49$}}$ (0.005) & $+0.44^{\mbox{\scriptsize$+0.18$}}_{\mbox{\scriptsize$-0.21$}}$ (\ensuremath{<}0.001) & $+0.020^{\mbox{\scriptsize$+0.009$}}_{\mbox{\scriptsize$-0.011$}}$ \\
\bottomrule
\end{tabular}

\end{table}

\begin{table}[h]
\centering
\caption{\textbf{Finetuning, magnitude pruning and DeiT-B.} Differences to Dense for the finetuning control and magnitude pruning (100 classes) and for DeiT-B (20 classes). Each group is matched among its own models, so differences to Dense vary slightly between groups (\cref{app:matched}). Size and core in points of the nodes, core/size as a ratio; $x^{+a}_{-b}$: 95\% interval from $x-b$ to $x+a$; smaller: classes whose circuit is smaller than Dense's.}
\label{tab:replications}
\footnotesize
\setlength{\tabcolsep}{2.5pt}
\begin{tabular}{llcccc}
\toprule
Comparison & Model & Size & Smaller & Core & Core/size \\
\midrule
Finetuning control & FT-Dense & $+0.24^{\mbox{\scriptsize$+0.25$}}_{\mbox{\scriptsize$-0.23$}}$ & 36/100 & $-0.01^{\mbox{\scriptsize$+0.24$}}_{\mbox{\scriptsize$-0.32$}}$ & $-0.004^{\mbox{\scriptsize$+0.013$}}_{\mbox{\scriptsize$-0.019$}}$ \\
 & Wanda 70\% (CE) & $+0.83^{\mbox{\scriptsize$+0.23$}}_{\mbox{\scriptsize$-0.24$}}$ & 30/100 & $+0.28^{\mbox{\scriptsize$+0.09$}}_{\mbox{\scriptsize$-0.14$}}$ & $+0.008^{\mbox{\scriptsize$+0.004$}}_{\mbox{\scriptsize$-0.007$}}$ \\
\midrule
Magnitude vs Wanda & Wanda 70\% (CE) & $+0.82^{\mbox{\scriptsize$+0.24$}}_{\mbox{\scriptsize$-0.25$}}$ & 28/100 & $+0.26^{\mbox{\scriptsize$+0.09$}}_{\mbox{\scriptsize$-0.13$}}$ & $+0.007^{\mbox{\scriptsize$+0.004$}}_{\mbox{\scriptsize$-0.007$}}$ \\
 & Magnitude 70\% (CE) & $+2.12^{\mbox{\scriptsize$+0.30$}}_{\mbox{\scriptsize$-0.29$}}$ & 8/100 & $+0.48^{\mbox{\scriptsize$+0.15$}}_{\mbox{\scriptsize$-0.17$}}$ & $+0.006^{\mbox{\scriptsize$+0.005$}}_{\mbox{\scriptsize$-0.009$}}$ \\
\midrule
DeiT-B finetuning control & FT-Dense & $+0.47^{\mbox{\scriptsize$+0.86$}}_{\mbox{\scriptsize$-0.75$}}$ & 10/20 & $+0.35^{\mbox{\scriptsize$+0.41$}}_{\mbox{\scriptsize$-0.36$}}$ & $+0.015^{\mbox{\scriptsize$+0.015$}}_{\mbox{\scriptsize$-0.016$}}$ \\
 & Wanda 70\% (CE) & $+1.57^{\mbox{\scriptsize$+0.67$}}_{\mbox{\scriptsize$-0.64$}}$ & 2/20 & $+0.19^{\mbox{\scriptsize$+0.23$}}_{\mbox{\scriptsize$-0.14$}}$ & $-0.004^{\mbox{\scriptsize$+0.010$}}_{\mbox{\scriptsize$-0.008$}}$ \\
\midrule
DeiT-B magnitude vs Wanda & Wanda 70\% (CE) & $+1.46^{\mbox{\scriptsize$+0.60$}}_{\mbox{\scriptsize$-0.58$}}$ & 2/20 & $+0.17^{\mbox{\scriptsize$+0.22$}}_{\mbox{\scriptsize$-0.17$}}$ & $-0.004^{\mbox{\scriptsize$+0.010$}}_{\mbox{\scriptsize$-0.010$}}$ \\
 & Magnitude 70\% (CE) & $+3.75^{\mbox{\scriptsize$+0.66$}}_{\mbox{\scriptsize$-0.62$}}$ & 0/20 & $+0.93^{\mbox{\scriptsize$+0.36$}}_{\mbox{\scriptsize$-0.07$}}$ & $+0.013^{\mbox{\scriptsize$+0.009$}}_{\mbox{\scriptsize$-0.006$}}$ \\
\midrule
DeiT-B & Wanda 70\% (KD) & $+1.25^{\mbox{\scriptsize$+0.63$}}_{\mbox{\scriptsize$-0.65$}}$ & 4/20 & $+0.06^{\mbox{\scriptsize$+0.15$}}_{\mbox{\scriptsize$-0.13$}}$ & $-0.008^{\mbox{\scriptsize$+0.007$}}_{\mbox{\scriptsize$-0.008$}}$ \\
 & Wanda 90\% (KD) & $+1.09^{\mbox{\scriptsize$+1.19$}}_{\mbox{\scriptsize$-1.28$}}$ & 8/20 & $+0.38^{\mbox{\scriptsize$+0.42$}}_{\mbox{\scriptsize$-0.33$}}$ & $+0.011^{\mbox{\scriptsize$+0.014$}}_{\mbox{\scriptsize$-0.014$}}$ \\
 & Wanda 90\% (CE) & $+1.07^{\mbox{\scriptsize$+1.47$}}_{\mbox{\scriptsize$-1.52$}}$ & 5/20 & $+0.31^{\mbox{\scriptsize$+0.42$}}_{\mbox{\scriptsize$-0.33$}}$ & $+0.007^{\mbox{\scriptsize$+0.015$}}_{\mbox{\scriptsize$-0.014$}}$ \\
\bottomrule
\end{tabular}

\end{table}

\begin{table}[h]
\centering
\caption{\textbf{Checks of the size comparison} (100 classes). Size difference to Dense when matching raw agreement instead of $\kappa$, and when selecting sizes on one half of each class's images and evaluating on the other ($\kappa$ as a difference); $x^{+a}_{-b}$: 95\% interval from $x-b$ to $x+a$.}
\label{tab:matched_checks}
\footnotesize
\begin{tabular}{lccc}
\toprule
 & Matched agreement & \multicolumn{2}{c}{Held-out (repeated splits)} \\
\cmidrule(lr){2-2}\cmidrule(lr){3-4}
Model & Size & Size & $\kappa$ \\
\midrule
Wanda 70\% (KD) & $+0.29^{\mbox{\scriptsize$+0.38$}}_{\mbox{\scriptsize$-0.58$}}$ & $+0.48^{\mbox{\scriptsize$+0.18$}}_{\mbox{\scriptsize$-0.18$}}$ & $-0.007^{\mbox{\scriptsize$+0.007$}}_{\mbox{\scriptsize$-0.008$}}$ \\
Wanda 70\% (CE) & $+0.51^{\mbox{\scriptsize$+0.46$}}_{\mbox{\scriptsize$-0.62$}}$ & $+0.62^{\mbox{\scriptsize$+0.22$}}_{\mbox{\scriptsize$-0.22$}}$ & $-0.009^{\mbox{\scriptsize$+0.008$}}_{\mbox{\scriptsize$-0.007$}}$ \\
Wanda 90\% (KD) & $-0.53^{\mbox{\scriptsize$+0.94$}}_{\mbox{\scriptsize$-1.07$}}$ & $+0.61^{\mbox{\scriptsize$+0.46$}}_{\mbox{\scriptsize$-0.48$}}$ & $-0.033^{\mbox{\scriptsize$+0.011$}}_{\mbox{\scriptsize$-0.011$}}$ \\
Wanda 90\% (CE) & $-1.02^{\mbox{\scriptsize$+1.08$}}_{\mbox{\scriptsize$-1.23$}}$ & $+0.43^{\mbox{\scriptsize$+0.47$}}_{\mbox{\scriptsize$-0.48$}}$ & $-0.042^{\mbox{\scriptsize$+0.012$}}_{\mbox{\scriptsize$-0.011$}}$ \\
\bottomrule
\end{tabular}

\end{table}

\cref{tab:lambda} repeats the size comparison at three sparsity penalties under the main protocol (20 stratified classes; Dense, Wanda 70\% (KD) and Wanda 90\% (KD) matched together at each $\lambda$). Dense circuits shrink from $\LamDenseSizeTwo\%$ to $\LamDenseSizeEight\%$ of the nodes as $\lambda$ grows fourfold, and the pruned models' circuits stay larger at every $\lambda$, with every interval above zero.

Matching by trimming also differs from retraining. A Dense circuit trained at $\lambda=8\times10^{-5}$ reaches the same $\kappa$ with $\TrimDDense$ points of the nodes ($\TrimLoDense$ to $\TrimHiDense$) fewer than the $\lambda=4\times10^{-5}$ circuit trimmed along its ranking, in the $\TrimNDense$ of 20 classes where the trimmed circuit can reach the trained $\kappa$. The difference is $\TrimDSeventy$ for Wanda 70\% (KD) ($\TrimNSeventy$ classes) and $\TrimDNinety$ for Wanda 90\% (KD) ($\TrimNNinety$ classes). Trimming thus overstates absolute circuit sizes for all three models alike, while the size differences to Dense are similar at $\lambda=4\times10^{-5}$ and $8\times10^{-5}$.

Giving each model its best of the three penalties, that is, its smallest circuit at the main per-class target $\kappa$ over the penalties that reach it, Dense needs $\BestDense\%$ of the nodes and the pruned circuits are larger by $\BestDiffSeventy$ (Wanda 70\% (KD); $\BestLoSeventy$ to $\BestHiSeventy$) and $\BestDiffNinety$ points (Wanda 90\% (KD); $\BestLoNinety$ to $\BestHiNinety$) on the same $\BestN$ classes. The direction holds, but only the 90\% interval excludes zero.

The matched $\kappa$ averages $\STargetKappa$, so \cref{tab:fixedkappa} repeats the comparison at fixed fidelity, on the classes whose trained circuits reach the level in every model. At $\kappa=0.6$ and $0.65$, all four pruned models have larger circuits with every interval above zero, and the differences of the 90\%-sparse models grow with $\kappa$. At $\kappa=0.7$ three of four are resolved (Wanda 70\% (CE) is not). At $\kappa=0.75$ and $0.8$ no pruned model's circuits are smaller; at $0.8$ ($\FkEightN$ classes) the 90\%-sparse models' circuits remain larger with intervals above zero, and the 70\%-sparse models' intervals include zero. At $\kappa=0.4$ the 90\%-sparse models' circuits are no larger than Dense's, with intervals that include zero.

\begin{table}[h]
\centering
\caption{\textbf{Size differences to Dense at fixed fidelity} (DeiT-III, 100 classes). Classes: those whose trained circuits reach $\kappa$ in every model, so the set shrinks with $\kappa$ and the Dense column is not a size--fidelity curve; Dense size in \% of the nodes; differences in points of the nodes with 95\% class-bootstrap intervals in brackets.}
\label{tab:fixedkappa}
\footnotesize
\setlength{\tabcolsep}{2.5pt}
\begin{tabular}{lcccccc}
\toprule
Fixed $\kappa$ & Classes & Dense (\%) & W70 KD & W70 CE & W90 KD & W90 CE \\
\midrule
0.4 & 98 & 14.2 & \ensuremath{+}0.26 [\ensuremath{+}0.03, \ensuremath{+}0.48] & \ensuremath{+}0.40 [\ensuremath{+}0.17, \ensuremath{+}0.65] & \ensuremath{-}0.03 [\ensuremath{-}0.48, \ensuremath{+}0.44] & \ensuremath{-}0.27 [\ensuremath{-}0.72, \ensuremath{+}0.20] \\
0.5 & 90 & 13.9 & \ensuremath{+}0.34 [\ensuremath{+}0.10, \ensuremath{+}0.55] & \ensuremath{+}0.41 [\ensuremath{+}0.16, \ensuremath{+}0.67] & \ensuremath{+}0.55 [\ensuremath{+}0.12, \ensuremath{+}0.96] & \ensuremath{+}0.30 [\ensuremath{-}0.16, \ensuremath{+}0.78] \\
0.55 & 84 & 13.6 & \ensuremath{+}0.34 [\ensuremath{+}0.11, \ensuremath{+}0.55] & \ensuremath{+}0.33 [\ensuremath{+}0.11, \ensuremath{+}0.56] & \ensuremath{+}0.78 [\ensuremath{+}0.34, \ensuremath{+}1.18] & \ensuremath{+}0.48 [\ensuremath{-}0.02, \ensuremath{+}0.97] \\
0.6 & 71 & 13.1 & \ensuremath{+}0.30 [\ensuremath{+}0.06, \ensuremath{+}0.52] & \ensuremath{+}0.33 [\ensuremath{+}0.08, \ensuremath{+}0.58] & \ensuremath{+}0.78 [\ensuremath{+}0.37, \ensuremath{+}1.19] & \ensuremath{+}0.64 [\ensuremath{+}0.18, \ensuremath{+}1.11] \\
0.65 & 65 & 13.1 & \ensuremath{+}0.30 [\ensuremath{+}0.10, \ensuremath{+}0.55] & \ensuremath{+}0.28 [\ensuremath{+}0.04, \ensuremath{+}0.54] & \ensuremath{+}0.82 [\ensuremath{+}0.38, \ensuremath{+}1.24] & \ensuremath{+}0.60 [\ensuremath{+}0.18, \ensuremath{+}1.06] \\
0.7 & 51 & 12.6 & \ensuremath{+}0.27 [\ensuremath{+}0.05, \ensuremath{+}0.52] & \ensuremath{+}0.23 [\ensuremath{-}0.06, \ensuremath{+}0.54] & \ensuremath{+}1.03 [\ensuremath{+}0.55, \ensuremath{+}1.54] & \ensuremath{+}0.87 [\ensuremath{+}0.34, \ensuremath{+}1.41] \\
0.75 & 43 & 12.3 & \ensuremath{+}0.39 [\ensuremath{+}0.11, \ensuremath{+}0.71] & \ensuremath{+}0.24 [\ensuremath{-}0.09, \ensuremath{+}0.60] & \ensuremath{+}1.21 [\ensuremath{+}0.63, \ensuremath{+}1.79] & \ensuremath{+}1.17 [\ensuremath{+}0.59, \ensuremath{+}1.78] \\
0.8 & 24 & 11.3 & \ensuremath{+}0.02 [\ensuremath{-}0.30, \ensuremath{+}0.38] & \ensuremath{+}0.08 [\ensuremath{-}0.31, \ensuremath{+}0.49] & \ensuremath{+}0.72 [\ensuremath{+}0.06, \ensuremath{+}1.37] & \ensuremath{+}0.94 [\ensuremath{+}0.20, \ensuremath{+}1.68] \\
\bottomrule
\end{tabular}

\end{table}

\begin{table}[h]
\centering
\caption{\textbf{Size differences to Dense at three sparsity penalties} (DeiT-III, 20 stratified classes, matched $\kappa$). Dense size in \% of the nodes; differences in points of the nodes; $x^{+a}_{-b}$: 95\% interval from $x-b$ to $x+a$.}
\label{tab:lambda}
\footnotesize
\begin{tabular}{lccc}
\toprule
$\lambda$ & Dense size (\%) & Wanda 70\% (KD) & Wanda 90\% (KD) \\
\midrule
$2\times10^{-5}$ & 20.6 & $+1.29^{\mbox{\scriptsize$+0.60$}}_{\mbox{\scriptsize$-0.67$}}$ & $+4.01^{\mbox{\scriptsize$+1.15$}}_{\mbox{\scriptsize$-1.19$}}$ \\
$4\times10^{-5}$ & 15.1 & $+0.68^{\mbox{\scriptsize$+0.49$}}_{\mbox{\scriptsize$-0.49$}}$ & $+1.49^{\mbox{\scriptsize$+0.73$}}_{\mbox{\scriptsize$-0.78$}}$ \\
$8\times10^{-5}$ & 7.5 & $+0.68^{\mbox{\scriptsize$+0.35$}}_{\mbox{\scriptsize$-0.33$}}$ & $+1.35^{\mbox{\scriptsize$+0.68$}}_{\mbox{\scriptsize$-0.63$}}$ \\
\bottomrule
\end{tabular}

\end{table}

\section{Units needed to decode a class}
\label{app:probe}
\begin{figure}[h]
\centering
\includegraphics[width=0.75\linewidth]{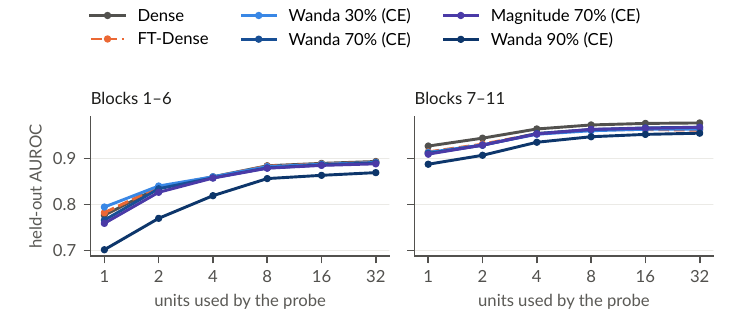}
\caption{\textbf{Units needed to decode a class.} Held-out AUROC of a logistic-regression probe on the $k$ most class-discriminative MLP units of a block, averaged over 100 classes and over early (left) or late (right) blocks.}
\label{fig:probe}
\end{figure}

Pruning could make circuits larger by spreading each class over more units. Sparse probes test this. They use the CLS-token activations of each block's MLP hidden units on the 5{,}000 validation images of the 100 classes. For each class and block, units are ranked by their single-unit AUROC on one random half of the images; a logistic-regression probe is then fitted to the top $k$ units on the same half and scored by its AUROC (the class against the other 99) on the other half, and the halves are swapped. Early blocks are blocks 1 to 6 and late blocks 7 to 11 (block 0 is not included). Sparse probes find no such spreading up to 70\% sparsity (\cref{fig:probe}). In early blocks the single most informative unit is slightly weaker after pruning (probe AUROC $\ProbeEarlyOneWandaSeventy$ for Wanda 70\% and $\ProbeEarlyOneMagSeventy$ for magnitude 70\%, compared with $\ProbeEarlyOneFtDense$ for FT-Dense), but eight units decode the class as well as before ($\ProbeEarlyEightWandaSeventy$ against $\ProbeEarlyEightFtDense$), and late blocks are unchanged. At 90\% sparsity early blocks lose some class information: the best single unit falls to $\ProbeEarlyOneWandaNinety$ and eight units reach $\ProbeEarlyEightWandaNinety$, while late blocks change little ($\ProbeLateEightWandaNinety$ against $\ProbeLateEightFtDense$).

\section{Noise in per-class accuracy}
\label{app:noise}
With 50 validation images per class, a per-class accuracy change is noisy, and \citet{hooker2019forget} trained 30 models per setting to separate real class effects from noise. We instead measure how reliable the per-class drops are. Splitting each class's images into halves and correlating the drops between halves gives a split-half reliability of $\DropRelWandaNinety$ at 90\% sparsity and $\DropRelWandaSeventy$ at 70\%. The class differences at 90\% are thus mostly real, and the correlations of \cref{sec:damage} are attenuated by noise rather than produced by it. Wanda and magnitude pruning damage largely the same classes (Spearman $\rho=\DropAgreeNinetyAll$ over all classes at 90\% sparsity). At 70\% the differences are small: finetuning alone, without pruning, changes per-class accuracy about as much (standard deviation $\DropSdFtVsDense$ points, reliability $\DropRelFtVsDense$) as 70\% pruning does on top of it ($\DropSdWandaSeventy$ points).

\section{Class accuracy loss and known factors}
\label{app:damage}
\cref{tab:damage_controls} compares dense circuit size with the factors that prior work links to disparate pruning damage. Per class we compute, on the dense model and its 50 validation images, the mean margin $z_c-\max_{j\neq c}z_j$ and the mean gradient norm $\|\nabla_\theta\,\mathrm{CE}(x,c)\|_2$~\citep{tran2022pruning}, the mean confidence~\citep{iofinova2023bias}, and, following \citet{paganini2020prune}, the number of training images and the mean entropy of the grayscale pixel histogram. Here, accuracy loss is measured relative to FT-Dense for CE-recovered models and relative to Dense otherwise. Partial correlations are Spearman correlations of the residual ranks after regressing out the ranks of the controls; controls that are constant over the classes are dropped. Circuit size correlates strongly ($|\rho|\approx0.7$) with the margin, the gradient norm and the confidence, yet it is more strongly associated with accuracy loss than any of these factors alone (\cref{tab:damage_controls}). At 90\% sparsity, magnitude pruning keeps $\CircKeptMinMagNinety$--$\CircKeptMaxMagNinety\%$ of the weights between the nodes of a class's dense circuit, almost the same share for every class, whereas Wanda keeps $\CircKeptMinWandaNinety$--$\CircKeptMaxWandaNinety\%$, less for larger circuits (Spearman $\rho=\CircKeptSizeRhoWandaNinety$ with circuit size).

To test prediction on classes not used for fitting, we compare linear models on ranks (fitted within each fold) of the five factors and the accuracy before pruning, with and without dense circuit size, by the Spearman correlation between out-of-fold predictions and observed accuracy loss (10-fold cross-validation over classes, repeated 20 times). Adding circuit size raises this correlation, with an interval above zero, for magnitude pruning to 90\% and one-shot to 70\%, but not resolvably for the three Wanda models (\cref{tab:prediction}; class-bootstrap 95\% intervals); holding out whole semantic groups gives the same picture.

\begin{table}[h]
\centering
\caption{\textbf{Dense circuit size and class accuracy loss.} Size: Spearman $\rho$ with the accuracy lost; best known factor: its strongest single correlate among the five factors and the accuracy before pruning; $|$accuracy, $|$all: partial $\rho$ controlling for accuracy before pruning or for all factors ($p$).}
\label{tab:damage_controls}
\footnotesize
\setlength{\tabcolsep}{2.5pt}
\begin{tabular}{lcccccc}
\toprule
 & \multicolumn{4}{c}{Main protocol (100 classes)} & \multicolumn{2}{c}{Class-sorted (100 classes)} \\
\cmidrule(lr){2-5}\cmidrule(lr){6-7}
Pruned model & Size & Best known factor & $|$accuracy & $|$all & $|$accuracy & $|$all \\
\midrule
Wanda 90\% (CE) & \ensuremath{+}0.44 & grad.\ norm \ensuremath{+}0.37 & \ensuremath{+}0.47 ($<$0.001) & \ensuremath{+}0.41 ($<$0.001) & \ensuremath{+}0.30 (0.002) & \ensuremath{+}0.22 (0.035) \\
Magnitude 90\% (CE) & \ensuremath{+}0.54 & grad.\ norm \ensuremath{+}0.38 & \ensuremath{+}0.54 ($<$0.001) & \ensuremath{+}0.50 ($<$0.001) & \ensuremath{+}0.29 (0.004) & \ensuremath{+}0.20 (0.048) \\
Wanda 90\% (KD) & \ensuremath{+}0.40 & grad.\ norm \ensuremath{+}0.35 & \ensuremath{+}0.38 ($<$0.001) & \ensuremath{+}0.27 (0.009) & \ensuremath{+}0.25 (0.014) & \ensuremath{+}0.15 (0.156) \\
Wanda 70\%, one-shot & \ensuremath{+}0.23 & grad.\ norm \ensuremath{+}0.19 & \ensuremath{+}0.24 (0.019) & \ensuremath{+}0.19 (0.067) & \ensuremath{+}0.11 (0.295) & \ensuremath{+}0.07 (0.502) \\
Magnitude 70\%, one-shot & \ensuremath{+}0.48 & grad.\ norm \ensuremath{+}0.37 & \ensuremath{+}0.45 ($<$0.001) & \ensuremath{+}0.38 ($<$0.001) & \ensuremath{+}0.28 (0.004) & \ensuremath{+}0.20 (0.050) \\
\bottomrule
\end{tabular}

\end{table}

\begin{table}[h]
\centering
\caption{\textbf{Adding dense circuit size helps predict accuracy loss on held-out classes only for magnitude pruning.} Change in the Spearman correlation between out-of-fold predictions and observed accuracy loss when circuit size is added to the known factors; $x^{+a}_{-b}$: 95\% class-bootstrap interval from $x-b$ to $x+a$.}
\label{tab:prediction}
\footnotesize
\begin{tabular}{lc}
\toprule
Model & Gain in held-out $\rho$ \\
\midrule
Wanda 90\% (CE) & $\ensuremath{+}0.08^{+0.10}_{-0.10}$ \\
Wanda 90\% (KD) & $\ensuremath{+}0.02^{+0.08}_{-0.08}$ \\
Wanda 70\%, one-shot & $\ensuremath{+}0.02^{+0.09}_{-0.10}$ \\
Magnitude 90\% (CE) & $\ensuremath{+}0.18^{+0.15}_{-0.14}$ \\
Magnitude 70\%, one-shot & $\ensuremath{+}0.11^{+0.10}_{-0.10}$ \\
\bottomrule
\end{tabular}

\end{table}

\section{Structural modularity per block}
\label{app:structure}
Each MLP block is a graph whose nodes are its 768 residual inputs, 3{,}072 hidden units and 768 residual outputs, with edge weights $|W|$; nodes whose weights were all pruned are dropped. Spectral clustering (100 $k$-means restarts) splits it into 12 clusters $V_i$, and the normalized cut is $\sum_i W(V_i,\bar V_i)/\mathrm{vol}(V_i)$. \cref{tab:structure} lists, for each MLP block graph, how far the normalized cut of the real weights lies below the mean of 20 copies shuffled within each matrix and, for pruned models in the last column, below 20 copies that shuffle only the surviving weights among the surviving positions. Larger values mean that the weights split into modules more cleanly than chance; the standard deviation over shuffles is about 0.002. Pruning produces more small clusters: clusters with fewer than 1\% of a block's nodes are $\TinyClusterShareDense\%$ of the clusters in Dense and $\TinyClusterShareWandaSeventy\%$ and $\TinyClusterShareWandaNinety\%$ in Wanda 70\% and 90\%, but in every model they hold at most $\TinyClusterNodesMax\%$ of the nodes.

\begin{table}[h]
\centering
\caption{Normalized cut below that of shuffled weights, per MLP block; ``Pattern'': mean over blocks against shuffles that keep the pruning pattern.}
\label{tab:structure}
\footnotesize
\setlength{\tabcolsep}{2.5pt}
\begin{tabular}{lcccccccccccccc}
\toprule
 & \multicolumn{13}{c}{Below full shuffle, block} & Pattern \\
\cmidrule(lr){2-14}
Model & 0 & 1 & 2 & 3 & 4 & 5 & 6 & 7 & 8 & 9 & 10 & 11 & Mean & Mean \\
\midrule
Dense & 0.45 & 0.45 & 0.45 & 0.37 & 0.24 & 0.28 & 0.26 & 0.21 & 0.10 & 0.18 & 0.19 & 0.12 & 0.27 & -- \\
FT-Dense & 0.45 & 0.45 & 0.44 & 0.37 & 0.24 & 0.28 & 0.25 & 0.20 & 0.08 & 0.17 & 0.19 & 0.09 & 0.27 & -- \\
Wanda 10\% (CE) & 0.45 & 0.46 & 0.44 & 0.36 & 0.26 & 0.29 & 0.24 & 0.22 & 0.10 & 0.15 & 0.17 & 0.13 & 0.27 & 0.33 \\
Wanda 30\% (CE) & 0.30 & 0.51 & 0.38 & 0.31 & 0.26 & 0.29 & 0.28 & 0.27 & 0.05 & 0.16 & 0.24 & 0.20 & 0.27 & 0.33 \\
Wanda 50\% (CE) & 0.60 & 0.47 & 0.52 & 0.44 & 0.40 & 0.30 & 0.32 & 0.33 & -0.02 & 0.24 & 0.41 & 0.22 & 0.35 & 0.34 \\
Wanda 70\% (CE) & 0.66 & 0.77 & 0.83 & 0.77 & 0.53 & 0.47 & 0.42 & 0.45 & 0.06 & 0.30 & 0.75 & 0.50 & 0.54 & 0.33 \\
Magnitude 70\% (CE) & 0.61 & 0.61 & 0.82 & 0.68 & 0.52 & 0.48 & 0.43 & 0.41 & 0.18 & 0.28 & 0.39 & 0.35 & 0.48 & 0.39 \\
Wanda 90\% (CE) & 1.47 & 1.32 & 1.19 & 1.77 & 0.85 & 1.16 & 0.70 & 0.84 & 0.22 & 0.65 & 1.11 & 1.08 & 1.03 & 0.48 \\
Wanda 70\% (KD) & 0.72 & 0.77 & 0.82 & 0.75 & 0.57 & 0.49 & 0.42 & 0.45 & 0.06 & 0.25 & 0.73 & 0.51 & 0.55 & 0.33 \\
Wanda 90\% (KD) & 1.32 & 1.30 & 1.25 & 1.78 & 0.85 & 1.16 & 0.81 & 0.84 & 0.23 & 0.75 & 1.09 & 1.05 & 1.03 & 0.47 \\
Wanda 80\% (CE) & 0.82 & 0.94 & 0.84 & 1.07 & 0.53 & 0.69 & 0.54 & 0.57 & 0.15 & 0.46 & 0.88 & 0.72 & 0.68 & 0.36 \\
Wanda 80\% (KD) & 0.79 & 0.88 & 0.84 & 1.08 & 0.54 & 0.68 & 0.53 & 0.57 & 0.15 & 0.45 & 0.89 & 0.71 & 0.68 & 0.35 \\
\bottomrule
\end{tabular}

\end{table}

\section{Circuit classes}
\label{app:classes}

\cref{tab:classes} lists the 100 classes by semantic group; the 20 starred classes are those DeiT-B uses.

\begin{table}[h]
\centering
\caption{The 100 classes by semantic group. Neurons and the accuracy-loss correlations use all 100. The circuit comparisons use all 100 in DeiT-III and the 20 starred classes in DeiT-B.}
\label{tab:classes}
\footnotesize
\begin{tabular}{lp{0.74\linewidth}}
\toprule
Group & Classes (index; $^\ast$ = the 20 classes used for DeiT-B) \\
\midrule
mammals & golden retriever (207), tabby cat (281), timber wolf (269), red fox (277), giant panda (388), african elephant (386), zebra (340$^\ast$), lion (291), brown bear (294), chimpanzee (367), arctic wolf (270), hedgehog (334), angora rabbit (332), killer whale (148$^\ast$), flying fox (273) \\
birds & bald eagle (22$^\ast$), flamingo (130), king penguin (145), peacock (84), great grey owl (24), hummingbird (94), toucan (96$^\ast$), rooster (7) \\
reptiles amphibians fish & king cobra (63), chameleon (44), loggerhead turtle (33$^\ast$), tree frog (31), american alligator (50$^\ast$), anemone fish (393), great white shark (2) \\
insects & monarch butterfly (323), ladybug (301), dragonfly (319), bee (309$^\ast$), scorpion (71) \\
vehicles & sports car (817), school bus (779), fire engine (555), mountain bike (671), schooner (780$^\ast$), airliner (404), warplane (895$^\ast$), steam locomotive (820), canoe (472), tractor (866) \\
food & pizza (963), cheeseburger (933), broccoli (937), banana (954), strawberry (949), ice cream (928$^\ast$), pretzel (932), espresso (967$^\ast$), mushroom (947), ear of corn (987) \\
scenes nature & volcano (980), coral reef (973), cliff (972), lakeside (975$^\ast$), sandbar (977$^\ast$), valley (979), seashore (978), alp (970) \\
structures & church (497), castle (483$^\ast$), beacon (437), suspension bridge (839), dam (525), triumphal arch (873), barn (425) \\
household & table lamp (532), grand piano (579$^\ast$), bookcase (453), analog clock (409), coffee mug (504$^\ast$), candle (470), rocking chair (765), television (851) \\
tools instruments & acoustic guitar (402), violin (889), hammer (587), chain saw (491), microscope (660), binoculars (447), stethoscope (823$^\ast$) \\
clothing accessories & running shoe (770$^\ast$), cowboy hat (515), sunglass (837), bow tie (457), backpack (414) \\
electronics & laptop (620), cellular telephone (487), joystick (613), traffic light (920), radio telescope (755$^\ast$) \\
abstract texture & spider web (815), jigsaw puzzle (611), maze (648$^\ast$), chain link fence (489), honeycomb (599) \\
\bottomrule
\end{tabular}

\end{table}

\section{Agreement between measures}
\label{app:crosslevel}
\cref{tab:crosslevel} gives the Spearman correlations between per-class changes from FT-Dense to Wanda 70\% (CE) on the 100 classes, with 95\% intervals from resampling classes, for the cross-measure test of \cref{sec:recovery}. Neurons and SAE features are scored by discriminability and label entropy (block 11), circuits of our protocol by their nodes and by their weights as a share of the model's surviving weights (at the trained $\lambda$, before fidelity matching), and attributions by the maps' gain over random maps in insertion and in deletion, averaged over the 50 validation images of each class.

A pair counts as weakly related only if its interval lies inside $\pm0.3$; $\XlNInside$ of the $\XlNPairs$ pairs do, $\XlCircInside$ of the $\XlNCircPairs$ that involve circuits. Every interval of a pair involving circuits lies within $\pm\XlCircCIBound$. The intervals are pointwise, not corrected for the $\XlNPairs$ pairs. The per-class changes are reliable: splitting each class's images in half gives a reliability of $\XlRelIns$ for the insertion gain and $\XlRelDel$ for the deletion gain, and two extractions with different image orders ($\XlRelNRepeat$ classes) give $\XlRelCirc$ for circuit size. Corrected for both, the largest circuit--attribution correlation rises only from $\XlCircAttrRho$ to $\XlCircAttrRhoCorr$.

With circuit size at matched fidelity instead (Wanda 70\% CE and FT-Dense matched together with Dense), no interval of the six pairs with the other measures excludes zero ($|\rho|\le\XlMatchedMaxRho$).

\begin{table}[h]
\centering
\caption{\textbf{Cross-measure test} (100 classes): Spearman correlation between per-class changes from FT-Dense to Wanda 70\% (CE) of two measures; $x^{+a}_{-b}$: pointwise 95\% interval from $x-b$ to $x+a$. Attribution ins./del.: gain of the maps over random maps in insertion and deletion.}
\label{tab:crosslevel}
\footnotesize
\begin{tabular}{llc}
\toprule
Measure & Measure & Spearman $\rho$ \\
\midrule
neuron discrim. & SAE discrim. & $+0.21^{\mbox{\scriptsize$+0.19$}}_{\mbox{\scriptsize$-0.20$}}$ \\
neuron entropy & SAE discrim. & $+0.03^{\mbox{\scriptsize$+0.21$}}_{\mbox{\scriptsize$-0.20$}}$ \\
neuron discrim. & SAE entropy & $-0.13^{\mbox{\scriptsize$+0.20$}}_{\mbox{\scriptsize$-0.20$}}$ \\
neuron entropy & SAE entropy & $+0.01^{\mbox{\scriptsize$+0.20$}}_{\mbox{\scriptsize$-0.19$}}$ \\
neuron discrim. & circuit weights & $-0.09^{\mbox{\scriptsize$+0.22$}}_{\mbox{\scriptsize$-0.20$}}$ \\
neuron entropy & circuit weights & $+0.05^{\mbox{\scriptsize$+0.21$}}_{\mbox{\scriptsize$-0.22$}}$ \\
SAE discrim. & circuit weights & $-0.03^{\mbox{\scriptsize$+0.21$}}_{\mbox{\scriptsize$-0.19$}}$ \\
SAE entropy & circuit weights & $+0.14^{\mbox{\scriptsize$+0.20$}}_{\mbox{\scriptsize$-0.21$}}$ \\
neuron discrim. & circuit nodes & $-0.14^{\mbox{\scriptsize$+0.21$}}_{\mbox{\scriptsize$-0.20$}}$ \\
neuron entropy & circuit nodes & $+0.14^{\mbox{\scriptsize$+0.20$}}_{\mbox{\scriptsize$-0.22$}}$ \\
SAE discrim. & circuit nodes & $+0.00^{\mbox{\scriptsize$+0.18$}}_{\mbox{\scriptsize$-0.21$}}$ \\
SAE entropy & circuit nodes & $+0.05^{\mbox{\scriptsize$+0.22$}}_{\mbox{\scriptsize$-0.20$}}$ \\
neuron discrim. & attribution ins. & $+0.27^{\mbox{\scriptsize$+0.16$}}_{\mbox{\scriptsize$-0.19$}}$ \\
neuron entropy & attribution ins. & $-0.03^{\mbox{\scriptsize$+0.20$}}_{\mbox{\scriptsize$-0.21$}}$ \\
SAE discrim. & attribution ins. & $-0.08^{\mbox{\scriptsize$+0.20$}}_{\mbox{\scriptsize$-0.20$}}$ \\
SAE entropy & attribution ins. & $+0.21^{\mbox{\scriptsize$+0.18$}}_{\mbox{\scriptsize$-0.20$}}$ \\
circuit weights & attribution ins. & $-0.15^{\mbox{\scriptsize$+0.20$}}_{\mbox{\scriptsize$-0.18$}}$ \\
circuit nodes & attribution ins. & $-0.06^{\mbox{\scriptsize$+0.18$}}_{\mbox{\scriptsize$-0.19$}}$ \\
neuron discrim. & attribution del. & $-0.06^{\mbox{\scriptsize$+0.18$}}_{\mbox{\scriptsize$-0.17$}}$ \\
neuron entropy & attribution del. & $+0.04^{\mbox{\scriptsize$+0.20$}}_{\mbox{\scriptsize$-0.19$}}$ \\
SAE discrim. & attribution del. & $+0.11^{\mbox{\scriptsize$+0.19$}}_{\mbox{\scriptsize$-0.20$}}$ \\
SAE entropy & attribution del. & $-0.18^{\mbox{\scriptsize$+0.20$}}_{\mbox{\scriptsize$-0.19$}}$ \\
circuit weights & attribution del. & $+0.09^{\mbox{\scriptsize$+0.18$}}_{\mbox{\scriptsize$-0.20$}}$ \\
circuit nodes & attribution del. & $-0.04^{\mbox{\scriptsize$+0.19$}}_{\mbox{\scriptsize$-0.19$}}$ \\
\bottomrule
\end{tabular}

\end{table}

\section{SAE training and quality}
\label{app:sae}
All SAEs use BatchTopK with $k=128$, $32\times$ expansion (24{,}576 latents), L1 coefficient $10^{-3}$, AuxK penalty $1/32$ on 512 auxiliary latents, learning rate $3\times10^{-4}$ with 500 warm-up steps, batch size 32, and three epochs (120{,}108 steps), on the unnormalized block-11 CLS token, and all feature metrics are computed on the CLS token. \cref{tab:sae_dense} reports the configuration sweep used to select $k$ and the expansion factor; the sweep runs are separate trainings from the SAEs in \cref{tab:sae_quality}.

FVE is the fraction of variance explained, $1-\mathrm{SS}_{\mathrm{res}}/\mathrm{SS}_{\mathrm{tot}}$ with the total sum of squares taken about the mean.

In the Dense sweep, training on the CLS token reconstructs far better than training on all or on patch tokens (FVE $\SaeSweepFveCls$ against $\SaeSweepFveAll$ and $\SaeSweepFvePatch$), FVE barely changes with the expansion factor at $k=128$ ($\SaeSweepFveExpMin$ to $\SaeSweepFveExpMax$ from $16\times$ to $64\times$), and $k=256$ raises it to $\SaeSweepFveKTwoFiftySix$ with twice the active latents per input.

The SAEs use block 11 because an SAE finds the most features there. Of the dense SAEs trained the same way on the CLS token after blocks 3, 5, 7, 9 and 11, the one at block 11 keeps the most live latents ($\SaeBlockLiveEleven\%$, against $\SaeBlockLiveThree$, $\SaeBlockLiveFive$, $\SaeBlockLiveSeven$ and $\SaeBlockLiveNine\%$). In blocks 3 and 5, on average $\SaeBlockLzeroThree$ and $\SaeBlockLzeroFive$ active latents reconstruct the CLS token almost perfectly (FVE $\SaeBlockFveThree$ and $\SaeBlockFveFive$), so an SAE there finds few features.

Most latents of the SAEs we use are dead ($\SaeDeadMin$--$\SaeDeadMax\%$, \cref{tab:sae_quality}), versus $\SaeSweepDeadAll\%$ when the Dense SAE is trained on all tokens, a limitation of training on the CLS token alone. The SAEs of pruned models also reconstruct their activations less well, so feature-level differences mix the model with the fit of its SAE.

A second SAE trained on the dense model gives the same discriminability as the first ($\SaeDiscDenseRetrain$ and $\SaeDiscDense$), so retraining the SAE does not account for the lower discriminability of finetuned and pruned models' features.

\begin{table}[h]
\centering
\caption{\textbf{Block-11 SAE reconstruction on the ImageNet validation set.} All SAEs share hyperparameters.}
\label{tab:sae_quality}
\footnotesize
\begin{tabular}{lcc}
\toprule
Model & FVE (\%) $\uparrow$ & Dead latents (\%) \\
\midrule
Dense & 99.25 & 85.9 \\
FT-Dense & 96.14 & 75.5 \\
Wanda 10\% (CE) & 95.96 & 75.0 \\
Wanda 30\% (CE) & 96.04 & 75.2 \\
Wanda 50\% (CE) & 95.39 & 73.0 \\
Wanda 70\% (CE) & 94.06 & 68.3 \\
\bottomrule
\end{tabular}

\end{table}

\begin{table}[h]
\centering
\caption{SAE sweep on Dense block-11 activations. Bold: the configuration used.}
\label{tab:sae_dense}
\footnotesize
\begin{tabular}{lccc}
\toprule
Tokens & $k$ & Expansion & FVE $\uparrow$ \\
\midrule
CLS & 128 & 16 & 0.9925 \\
\textbf{CLS} & \textbf{128} & \textbf{32} & \textbf{0.9926} \\
CLS & 128 & 64 & 0.9926 \\
CLS & 32 & 32 & 0.9869 \\
CLS & 64 & 32 & 0.9898 \\
CLS & 256 & 32 & 0.9949 \\
All & 128 & 32 & 0.9476 \\
Patch & 128 & 32 & 0.9474 \\
\bottomrule
\end{tabular}

\end{table}

\section{Circuit extraction}
\label{app:circuits}
\begin{figure*}[h]
\centering
\includegraphics[width=6.8in]{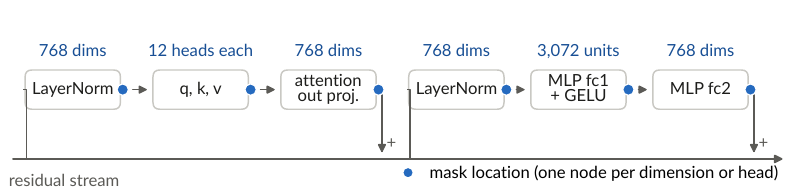}
\caption{\textbf{Circuit nodes in one transformer block.} A mask sits after each of the eight operations (blue dots), with one node per dimension, per hidden unit, or per attention head for the query, key and value; 6{,}180 nodes per block, 74{,}160 in the 12 blocks.}
\label{fig:nodes}
\end{figure*}
\paragraph{Circuit masks.}
Masks sit at eight locations per block (both LayerNorm outputs, the query, key, and value of each head, the attention output, the MLP hidden units, and the MLP output), 74{,}160 nodes in total. They are trained with a sigmoid surrogate gradient to minimize the task loss plus $\lambda$ times the expected number of active nodes, $\sum_i\sigma(m_i)$ for mask logits $m_i$ ($\lambda$ and the schedule follow below). Mask logits $m$ are initialized from $\mathcal{N}(0.5,0.01^2)$ and clamped to $[-1,1]$ after each step. The forward pass uses $\hat m=\mathbb{1}[m>0]$ and the backward pass the surrogate gradient $\partial\hat m/\partial m=\tau\,\sigma(\tau m)(1-\sigma(\tau m))$ with $\tau=1$. DeiT-III fuses the query, key, and value projections; we slice the fused output and mask each head's query, key, and value separately. Masks are shared across tokens and images.

\paragraph{Optimization.} 
Masks are optimized for 45 epochs with AdamW (learning rate $10^{-2}$, weight decay $0.01$), a step schedule that multiplies the learning rate by $0.2$ every 15 epochs, and gradient clipping at norm 1. The learning rate, the number of epochs and $\lambda$ come from a cost-aware Bayesian search (CARBS~\citep{fetterman2023tune}) on Dense under the 1000-way objective (defined below); a smaller search on Wanda 70\% (CE) favored learning rates that include ours, and the step schedule was set by hand.

All models use $\lambda=4\times10^{-5}$ (other penalties: \cref{app:matched}).

Random baselines permute each learned mask within its location, which keeps the number of active nodes per location, and are evaluated the same way as the circuit.

\paragraph{Objectives.} \emph{1000-way}: cross-entropy of the 1000-way prediction against the labels, on all images of class $c$ and as many images of other classes, with masked nodes replaced by their mean over all tokens. 

\emph{Class vs rest}: two-way cross-entropy of $[\max_{j\neq c} z_j,\ z_c]$ against whether the image belongs to $c$; the maximum, not the log-sum-exp of the other logits, makes the sign of the margin the argmax decision. 

\emph{Faithful} (our protocol): the same two-way cross-entropy against the full model's own decision ($\arg\max_j z^{\mathrm{model}}_j = c$), with masked nodes replaced by their mean at the same token position, and with every LayerNorm dividing by $\max(s_{\mathrm{clean}},s)$, where $s_{\mathrm{clean}}$ is the standard deviation the full model computes on the same input (one clean forward pass per batch) and $s$ the current one. Dividing by $s_{\mathrm{clean}}$ alone overflows (\cref{app:extraction}); training steps with a non-finite loss are skipped.

\paragraph{Training order.} Each class's training set contains its training images and as many images of other classes. In our protocol the set is shuffled once (fixed seed) and presented in that order in every epoch. A class-sorted order presents all images of the class first, then the others, in the same order every epoch; \cref{app:extraction} compares the two.

Circuits are not bit-reproducible on GPU: two identical runs differ in about 3\% of the node decisions.

\begin{figure}[t]
\centering
\includegraphics[width=3.25in]{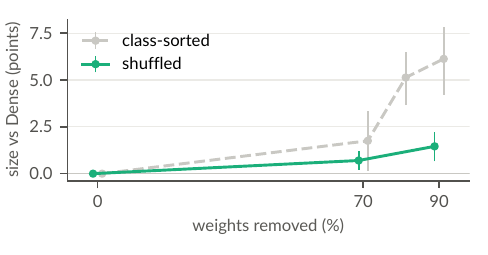}
\caption{\textbf{A class-sorted training order inflates the circuit-size difference.} Change in class-circuit size relative to the dense model at matched fidelity, for the distilled models, with circuits trained on class-sorted or shuffled images (20 classes; 95\% intervals).}
\label{fig:order}
\end{figure}

\paragraph{Classes and matching.} The 100 classes come from $\NClassGroups$ semantic groups; we chose them by hand to span the groups. Every DeiT-III model in the circuit comparisons has circuits for all 100 classes, and every DeiT-B model for the 20 starred classes (\cref{tab:classes}). Circuits are evaluated on all 50 validation images of the class and 50 images of other classes.

For every model and class we evaluate the circuits formed by the top-$k$ nodes of the learned mask logits for 37 values of $k$ between 10\% and 100\% of the trained circuit, on the validation images and under the training conditions. Beyond the trained circuit the ranking is not informative (nodes the training switched off break the circuit again), so the matching stays inside it. For each class the target fidelity is the lowest $\kappa$ that any model's trained circuit reaches, and each model's circuit is the smallest one that reaches it.

For the held-out check, the same curves are evaluated per image, the target and each model's $k$ are chosen on one half of the images, and $\kappa$ is measured on the other. The means that replace removed nodes are computed once per model from the full validation set (per token position for our protocol) and are shared by every circuit, so the held-out check isolates the choice of circuit size while the ablation means stay fixed.

\paragraph{Restoration.} One-shot Wanda is calibrated on one training image per class, as for the distilled models. Damage is measured on the 50 validation images of each class. The trained circuits are learned on training images only.

A weight lies in a node set if its output node (MLP hidden unit, MLP output, attention head of the query, key, or value, or attention output) and its input node (LayerNorm output, MLP hidden unit, or value head) both do; the classifier is not included. The three other classes of each class are the next, fifth-next, and tenth-next in the class list.

The score-matched control bins the pruned weights of every layer into deciles of the pruning score and restores, per layer and decile, as many weights from outside the own circuit as the own condition restored there; where a decile held too few, the remainder came from the closest scores in the same layer. The fallback never crosses layers, so where a layer had fewer pruned weights outside the circuit than the own condition restored there, the control restored fewer weights; restored counts are recorded for every class. Damage is summed over classes before taking the share undone, so classes with more damage weigh more.

Because the matched control can fall short of its budget and fills short score deciles from neighboring scores, we also run an equal-budget check. It draws the own-circuit restoration, counts it in every (layer, pruning-score decile) stratum, caps each stratum at the pruned weights available outside the circuit, and then restores exactly that many weights at random inside the circuit (own) and outside it (control), three draws per class.

We also compare the own circuit with class-specific weight scores that need no circuit (\cref{tab:restoration_baselines}). Each score is computed on 256 training images of the class, never on the validation images used for evaluation, and restores the top-scoring pruned weights of the block layers with the same budget: a Wanda score with activation norms from the class's images; $|w\,\partial L_c/\partial w|$ in the dense model, with $L_c$ the cross-entropy toward the class on its images, or with a class-vs-rest loss that adds cross-entropy toward the true labels on as many images of other classes; and $-w\,\partial L_c/\partial w$ in the pruned model, with $w$ the weight's dense value, the first-order predicted drop of $L_c$ when a weight is restored, with the same two losses. A restoration counts as a repair only if, on two validation images of each other class, the share predicted as the class rises by at most one point and top-1 accuracy falls by at most one point (the specificity rule).

\section{Restoration results}
\label{app:restoration}
After one-shot Wanda pruning to 70\%, restoring the own circuit's weights raises mean class top-1 accuracy from $\ProtAccNoneWanda\%$ to $\ProtAccOwnWanda\%$. Random node sets undo at most $\ProtUndoRandMax\%$ of the damage; other classes' circuits help more, as expected if circuits share nodes. The score-matched control undoes the least because it draws only from outside the class's circuit, whereas random nodes can fall inside it.

The number of restored weights is on average $\ProtOwnFracWanda\%$ (Wanda) and $\ProtOwnFracMag\%$ (magnitude) of the own circuit's pruned weights, and $\ProtWeightsWanda\%$ and $\ProtWeightsMag\%$ of the 84.9 million weights of the blocks' linear layers. The score-matched control restored at least $\ProtMatchedFilled\%$ of the requested weights in every class. Resampling classes gives 95\% intervals of $\RestOwnLoWanda$--$\RestOwnHiWanda\%$ (Wanda) and $\RestOwnLoMag$--$\RestOwnHiMag\%$ (magnitude) for the share the own circuit undoes.

The equal-budget check restores at least $\ExactCoverMinWanda\%$ (Wanda) and $\ExactCoverMinMag\%$ (magnitude) of the requested budget. The own condition undoes $\ExactOwnWanda\%$ and $\ExactOwnMag\%$ of the summed damage, the control $\ExactCtlWanda\%$ and $\ExactCtlMag\%$. The own condition wins in $\ExactWinsWanda$ and $\ExactWinsMag$ of $\ExactNWanda$ classes, with a mean paired log-probability gain between $\ExactGainLoWanda$ and $\ExactGainHiWanda$ (Wanda) and between $\ExactGainLoMag$ and $\ExactGainHiMag$ (magnitude; class-bootstrap 95\% intervals). In DeiT-B the own condition undoes $\ExactOwnWandaDeitb\%$ against $\ExactCtlWandaDeitb\%$ (Wanda) and $\ExactOwnMagDeitb\%$ against $\ExactCtlMagDeitb\%$ (magnitude).

The pruned-model gradient scores fail the specificity rule of \cref{app:circuits}. They restore weights that push nearly every image toward the class, raising that share by $\BaseInvalidFpMin$ to $\BaseInvalidFpMax$ points. They also break the model for the class itself: in $\BaseBreakWanda$ (Wanda) and $\BaseBreakMag$ (magnitude) of the $\BaseBreakNWanda$ classes the class's own log-probability falls below its value after pruning, down to $-\BaseBreakLogpWanda$ and $-\BaseBreakLogpMag$ nats, which outweighs the other classes in the summed damage and makes the share undone negative. The own circuit is drawn again for this comparison, so its numbers differ slightly from those above.

\begin{table}[h]
\centering
\caption{\textbf{Class-specific weight scores against the own circuit} (100 classes, one-shot pruning to 70\%, same budget per class). Undone: \% of the summed damage; $\Delta$FP, $\Delta$top-1: change in the share of other classes' images predicted as the class and in their top-1 accuracy (points, relative to the pruned model); wins: classes where the own circuit undoes more; $^\dagger$: fails the specificity rule.}
\label{tab:restoration_baselines}
\footnotesize
\begin{tabular}{lrrrrrrrr}
\toprule
 & \multicolumn{4}{c}{Wanda 70\%, one-shot} & \multicolumn{4}{c}{Magnitude 70\%, one-shot} \\
\cmidrule(lr){2-5}\cmidrule(lr){6-9}
Restored weights & Undone & $\Delta$FP & $\Delta$top-1 & Wins & Undone & $\Delta$FP & $\Delta$top-1 & Wins \\
\midrule
Own circuit (paired) & 95 & \ensuremath{+}0.1 & \ensuremath{+}4.1 & -- & 82 & \ensuremath{+}0.1 & \ensuremath{+}8.7 & -- \\
Wanda score on class images & 65 & \ensuremath{+}0.0 & \ensuremath{+}2.5 & 92 & 55 & \ensuremath{+}0.0 & \ensuremath{+}10.4 & 96 \\
$|w\,\partial L_c/\partial w|$, dense & 39 & \ensuremath{+}0.1 & \ensuremath{+}6.4 & 89 & 66 & \ensuremath{+}0.1 & \ensuremath{+}11.0 & 73 \\
same, class vs rest & 34 & \ensuremath{+}0.1 & \ensuremath{+}6.6 & 91 & 66 & \ensuremath{+}0.0 & \ensuremath{+}11.8 & 76 \\
$-w\,\partial L_c/\partial w$, pruned & \ensuremath{-}40$^\dagger$ & \ensuremath{+}80.0 & \ensuremath{-}67.0 & 30 & \ensuremath{-}235$^\dagger$ & \ensuremath{+}60.1 & \ensuremath{-}60.0 & 48 \\
same, class vs rest & 21$^\dagger$ & \ensuremath{+}87.7 & \ensuremath{-}67.0 & 23 & \ensuremath{-}154$^\dagger$ & \ensuremath{+}70.2 & \ensuremath{-}59.9 & 39 \\
\bottomrule
\end{tabular}

\end{table}

\section{Extraction choices that change the comparison}
\label{app:extraction}

Five extraction choices can change an apparent difference in circuit size or sharing between dense and pruned models (\cref{tab:protocol}). For each, we give the failure it causes, the comparison that shows it, and the choice we use. The training order alone can make pruning to 90\% appear to enlarge class circuits by six points and triple their shared core.

\paragraph{Training order.} The order in which a class's training images are presented can decide which of several similarly faithful circuits the optimization finds. With class-sorted batches (all images of the class, then the other images, in the same order every epoch), the masks of Dense end smaller and less faithful: $\OrderLambdaSortedDense\%$ of the nodes at the trained $\lambda$ and $\OrderAgreeSortedDense\%$ agreement, compared with $\OrderLambdaShufDense\%$ and $\OrderAgreeShufDense\%$ with a shuffled order. The masks of the 70\%-sparse distilled model do the same, whereas the circuits of the 90\%-sparse models hardly change. At matched fidelity, the sorted order makes the distilled 90\% model's circuits $\OrderSortedDiffNinetyKd$ points larger than Dense's, versus $\OrigSizeDiffWandaNinetyKd$ with the shuffled order on the same 20 classes (\cref{fig:order}), and gives shared cores of $\OrderSortedCoreDense\%$ of the nodes in Dense and $\OrderSortedCoreNinetyKd\%$ in the distilled 90\% model.

On six of these classes, the difference between the distilled 90\% model and Dense is $\DiagNinetyOrig$ points with class-sorted batches, $\DiagNinetyRepeat$ with the same order and a new mask seed, and $\DiagNinetyShuf$ with the shuffled order, so the order, not the seed, sets it. A second shuffled order gives $\DiagNinetyShufTwo$ and $\DiagSeventyShufTwo$ points for the 90\% and 70\% distilled models, compared with $\DiagNinetyShuf$ and $\DiagSeventyShuf$ for the first. Two extractions with the same order share only a fraction of their nodes (Jaccard similarity of the node sets $\DiagJaccardRepeat$) while their sizes agree, so we compare sizes and shares, not node identities. We use one fixed shuffled order for every circuit.

\paragraph{Objective.} An objective that asks a circuit to do more than the class decision fills the shared core with machinery that every circuit needs. Circuits trained to predict the labels of all 1000 classes on a balanced set of class and other images must also keep what classifies the other images. This 1000-way label-trained extraction gives a shared core of $\CoreHundredDense\%$ of the dense model's nodes and $\CoreHundredWandaNinety\%$ at 90\% sparsity (100 classes), compared with $\SCoreDense\%$ and $\SCoreWandaNinety\%$ for our circuits, which reproduce only the class-vs-rest decision. The two protocols also differ in the target, the ablation means and the LayerNorm treatment, so the comparison bounds the effect of the objective rather than isolating it. We train class-vs-rest circuits.

\paragraph{Target.} Circuits trained to separate their class from the rest according to the labels can implement rules the model does not compute. Asked to map the images of an unrelated class to the logit of the target class (a \emph{swap control}), they reach $\SwapHeldoutDense\%$ held-out accuracy with $\SwapNodeDense\%$ of the dense model's nodes ($\SwapHeldoutWandaNinety\%$ in Wanda 90\%), while random relabelings, which have no rule, stay near chance ($\RndHeldoutDense\%$). Wanda 90\%'s label-trained circuits even separate their class better than the model does ($\BinFidWandaNinety\%$ against $\BinFullAccWandaNinety\%$). We train circuits on the model's own decisions, which rewards a circuit only for agreeing with the model and removes the incentive for both.

\paragraph{LayerNorm.} Ablating nodes lowers the variance of the residual stream, and LayerNorm scales it back up, so a few nodes can stand in for many~\citep{drori2026unfaithful}. With LayerNorm frozen to the unmasked model's scale, label-trained circuits lose $\BinFrozenDropDense$ (Dense), $\BinFrozenDropWandaSeventy$ (Wanda 70\%), and $\BinFrozenDropWandaNinety$ points (Wanda 90\%) of accuracy. Dense circuits rely on rescaling most, which biases size comparisons between models.

Freezing the scale exactly during training is unstable, because mean values can inflate a token's variance far beyond the clean one and the division then overflows; with the model's decisions as the target, $\CollapseVOneTotal$ class circuits collapsed to zero nodes ($\CollapseVOneDense$ of 20 for Dense, $\CollapseVOneWandaSeventyKd$ for Wanda 70\% (KD), $\CollapseVOneWandaSeventy$ for Wanda 70\% (CE)), which alone emptied the shared core of the dense and distilled models. We therefore use the LayerNorm clamp, which divides by the larger of the clean and the current scale and so forbids amplification without overflow. Clamped circuits thus take their LayerNorm scale from the full model and are not standalone subnetworks.

The choice matters for the size comparison. The curves below re-evaluate the clamped masks without the clamp, on a log-spaced grid of sizes over all nodes (not the 37 sizes of \cref{app:circuits}), with steps of about a third in node count. Along the whole ranked size curve of $\CurveN$ classes, matched at equal agreement, the distilled models' circuits are $\CurveClampSeventyKd$ (70\%) and $\CurveClampNinetyKd$ (90\%) points larger than the dense model's with the clamp (\CurveClampCISeventyKd{} and \CurveClampCINinetyKd{}), but $\CurveNativeSeventyKd$ and $\CurveNativeNinetyKd$ points smaller without the clamp (\CurveNativeCISeventyKd{} and \CurveNativeCINinetyKd{}; $\CurveNativeSmallerSeventyKd$ and $\CurveNativeSmallerNinetyKd$ of $\CurveN$ classes smaller). The size comparison therefore holds under the LayerNorm clamp and not without it. Without the clamp, the dense and 70\%-sparse circuits lose far more fidelity than the 90\%-sparse ones, so the pruned circuits look smaller; neither unclamped difference has an interval that excludes zero.

\paragraph{Fidelity.} At a fixed $\lambda$, circuits of models whose decisions are easier to reproduce stop at higher fidelity and overlap more, so we compare circuits at matched fidelity, as Gao~\etal\ do. Raw agreement is not enough for the match, because it counts agreement expected by chance, which is high when a model rarely predicts the class (\cref{sec:circuits}). Matching agreement makes the distilled 90\% model's circuits look smaller, $\SSizeAgWandaNinetyKd\%$ instead of $\SSizeWandaNinetyKd\%$ of the nodes. We match Cohen's $\kappa$.

\begin{table}[h]
\centering
\caption{\textbf{Five properties of the extraction change the comparison of dense and pruned circuits} (\cref{app:extraction}): the alternative tested, its effect on the comparison, and the choice we use. 20 of the 100 classes, except the objective and fidelity rows (all 100).}
\label{tab:protocol}
\footnotesize
\begin{tabular}{@{}>{\raggedright\arraybackslash}p{0.13\linewidth}>{\raggedright\arraybackslash}p{0.19\linewidth}>{\raggedright\arraybackslash}p{0.43\linewidth}>{\raggedright\arraybackslash}p{0.17\linewidth}@{}}
\toprule
Property & Alternative tested & Effect & Our choice \\
\midrule
Training order & class-sorted batches & Wanda 90\% (KD) circuits $\OrderSortedDiffNinetyKd$ points larger than Dense's (shuffled order: $\OrigSizeDiffWandaNinetyKd$) & one fixed shuffled order \\
Objective & 1000-way label-trained extraction & shared core $\CoreHundredDense\%\to\CoreHundredWandaNinety\%$ from Dense to Wanda 90\% (ours: $\SCoreDense\%\to\SCoreWandaNinety\%$) & class vs rest \\
Target & the labels & circuits learn a swap rule ($\SwapHeldoutDense\%$ held out) & the model's own decision \\
LayerNorm & free rescaling; the exact clean scale & freezing costs $\BinFrozenDropDense$ points; the exact scale collapses $\CollapseVOneTotal$ circuits & the LayerNorm clamp \\
Fidelity & raw agreement & Wanda 90\% (KD) circuits look smaller ($\SSizeAgWandaNinetyKd\%$ vs $\SSizeWandaNinetyKd\%$ of nodes) & matched Cohen's $\kappa$ \\
\bottomrule
\end{tabular}
\end{table}

\section{Neuron effects at blocks 2 and 7}
\label{app:blocks}
\begin{figure}[h]
\centering
\includegraphics[width=0.9\linewidth]{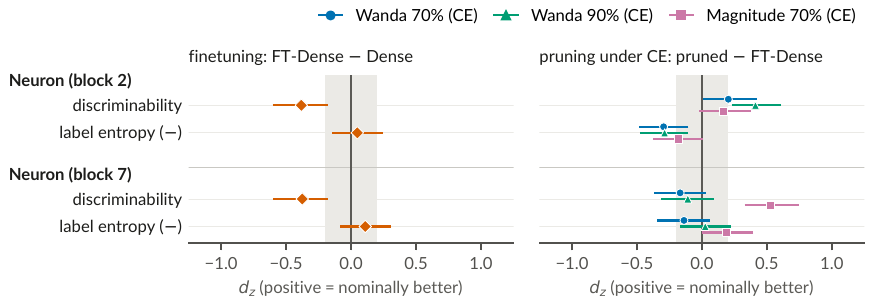}
\caption{Finetuning and sparsity effects on neuron metrics at blocks 2 and 7, as in \cref{fig:decomposition}.}
\label{fig:blocks}
\end{figure}

\cref{fig:blocks} repeats the neuron comparison of \cref{fig:decomposition} for blocks 2 and 7. Finetuning lowers unit discriminability in both blocks as in block 11, though less, and leaves label entropy nearly unchanged; pruning's effects there vary in sign across blocks and methods.

\section{Suppressing units and raw AUROC}
\label{app:suppressors}

Attribution patching ranks units by the magnitude of their effect on the class logit, so the selected set contains units whose activation lowers the class logit. For such a unit, activation is lower on images of the class, and its AUROC falls below 0.5. Averaging raw AUROC over the selected units mixes the two kinds. In block 11 the mean raw AUROC is $\NeuronRawDense$ for Dense, $\NeuronRawFtDense$ for FT-Dense, and $\NeuronRawWandaSeventy$ for Wanda 70\%. The discriminability $\max\{A,1-A\}$ used in the main text scores both kinds by how well they separate the class.

\section{Statistical tests}
\label{app:stats}

With 100 to 1{,}000 classes nearly every comparison would be significant in a paired test, so for neurons, SAE features, and attributions we base conclusions on effect sizes and their direction, not on $p$-values. For neurons and SAE features the paired unit is the class; attribution scores are per image, so their effect sizes and intervals use the 2{,}000 paired evaluation images, which come from fewer than 1{,}000 classes. Circuits of our protocol are compared across models on 100 classes (20 in DeiT-B), and their shared core is a property of the set of circuits rather than of single classes; there we report intervals from resampling classes, with the core recomputed in every resample, and paired permutation tests (\cref{app:matched}).

\section{Coverage and reproducibility}
\label{app:coverage}
\cref{tab:coverage} lists the measures computed for each model and, for circuits, the number of classes with circuits of the main protocol. Every result in the main text and every data figure is generated from the saved results by one script, and a second script checks the quoted values and stated comparisons it lists against them. Both scripts, the code for pruning, recovery, circuit extraction and evaluation, the class lists, seeds and the circuit masks will be released separately.

\begin{table}[h]
\centering
\caption{\textbf{Coverage.} Measures per model; circuits: classes with circuits of our protocol; neurons: classes with block-11 unit measures. One-shot 70\% models enter only the restoration test and the class-damage analysis, magnitude 90\% only the latter.}
\label{tab:coverage}
\footnotesize
\begin{tabular}{lccccccc}
\toprule
Model & Modularity gap & Circuits (classes) & Neurons (classes) & SAE & Attribution & Drift & Acc. kept \\
\midrule
Dense & \checkmark & 100 & 1000 & \checkmark & \checkmark & \checkmark & \checkmark \\
FT-Dense & \checkmark & 100 & 1000 & \checkmark & \checkmark & \checkmark & \checkmark \\
FT-Dense (gentle) &  &  & 100 &  & \checkmark & \checkmark & \checkmark \\
Wanda 10--50\% (CE) & \checkmark &  & 1000 & \checkmark & \checkmark & \checkmark & \checkmark \\
Wanda 70\% (CE) & \checkmark & 100 & 1000 & \checkmark & \checkmark & \checkmark & \checkmark \\
Wanda 70\% (BCE) &  &  & 100 &  & \checkmark & \checkmark & \checkmark \\
Wanda 70\% (KD) & \checkmark & 100 & 100 &  & \checkmark & \checkmark & \checkmark \\
Magnitude 70\% (CE) & \checkmark & 100 & 1000 &  & \checkmark & \checkmark & \checkmark \\
Wanda 90\% (CE) & \checkmark & 100 & 1000 &  & \checkmark & \checkmark & \checkmark \\
Wanda 90\% (KD) & \checkmark & 100 & 100 &  & \checkmark & \checkmark & \checkmark \\
DeiT-B Dense & \checkmark & 20 & 100 & \checkmark & \checkmark & \checkmark & \checkmark \\
DeiT-B FT-Dense & \checkmark & 20 & 100 & \checkmark & \checkmark & \checkmark & \checkmark \\
DeiT-B Wanda 70\% (CE) & \checkmark & 20 & 100 & \checkmark & \checkmark & \checkmark & \checkmark \\
DeiT-B Wanda 70\% (KD) & \checkmark & 20 & 100 &  & \checkmark & \checkmark & \checkmark \\
DeiT-B Magnitude 70\% (CE) & \checkmark & 20 & 100 &  & \checkmark & \checkmark & \checkmark \\
DeiT-B Wanda 90\% (KD) & \checkmark & 20 & 100 &  & \checkmark & \checkmark & \checkmark \\
DeiT-B Wanda 90\% (CE) & \checkmark & 20 & 100 &  & \checkmark & \checkmark & \checkmark \\
\bottomrule
\end{tabular}

\end{table}

\section{Attribution faithfulness}
\label{app:attribution}
\cref{tab:attribution} gives the deletion and insertion AUCs of transformer-attribution maps for every model, raw and divided by the clean-image confidence $q$, with uniform-random maps as a reference. Deletion replaces pixels by zeros, and insertion adds the image's pixels to a zero image, 224 pixels per step, in the order the map ranks them. The gain over random maps is computed per image from the raw AUCs. For insertion, the map's AUC minus that of a uniform-random map on the same image; for deletion, the random map's AUC minus the map's. Both are averaged over the 2{,}000 images, and higher means the map beats a random order. The insertion gain is $\AdvInsDense$ for Dense and $\AdvInsFtDense$ for FT-Dense, and distillation raises it by $\AdvInsDiffWandaSeventyKd$ at 70\% and $\AdvInsDiffWandaNinetyKd$ at 90\% sparsity. Insertion divided by $q$, by contrast, falls with $d_z=\DzDenseInsCalWandaSeventyKd$ and $\DzDenseInsCalWandaNinetyKd$ against Dense.

\begin{table}[h]
\centering
\caption{\textbf{Attribution faithfulness} on 2{,}000 validation images. $q$: clean-image probability of the predicted class; ``/$q$'': per-image AUC divided by $q$; ``rand.'': uniform-random maps.}
\label{tab:attribution}
\footnotesize
\setlength{\tabcolsep}{3.5pt}
\begin{tabular}{lccccccc}
\toprule
Model & Conf.\ $\bar q$ & Ins.\,$\uparrow$ & Del.\,$\downarrow$ & Ins./$q$\,$\uparrow$ & Del./$q$\,$\downarrow$ & Ins.\ (rand.) & Del.\ (rand.) \\
\midrule
Dense & 0.775 & 0.599 & 0.216 & 0.770 & 0.265 & 0.394 & 0.394 \\
FT-Dense & 0.889 & 0.708 & 0.210 & 0.790 & 0.223 & 0.384 & 0.385 \\
\midrule
Wanda 10\% (CE) & 0.891 & 0.702 & 0.206 & 0.783 & 0.220 & 0.383 & 0.384 \\
Wanda 30\% (CE) & 0.891 & 0.702 & 0.206 & 0.784 & 0.220 & 0.397 & 0.396 \\
Wanda 50\% (CE) & 0.886 & 0.706 & 0.205 & 0.790 & 0.221 & 0.388 & 0.389 \\
Wanda 70\% (CE) & 0.857 & 0.662 & 0.180 & 0.766 & 0.197 & 0.316 & 0.317 \\
Wanda 90\% (CE) & 0.591 & 0.354 & 0.084 & 0.577 & 0.131 & 0.093 & 0.093 \\
Magnitude 70\% (CE) & 0.862 & 0.665 & 0.171 & 0.764 & 0.187 & 0.299 & 0.299 \\
\midrule
Wanda 70\% (KD) & 0.748 & 0.557 & 0.184 & 0.738 & 0.231 & 0.301 & 0.301 \\
Wanda 90\% (KD) & 0.524 & 0.321 & 0.076 & 0.599 & 0.138 & 0.082 & 0.082 \\
\bottomrule
\end{tabular}

\end{table}

\section{Qualitative attributions}
\label{app:qualitative}
\begin{figure}[h]
\centering
\setlength\tabcolsep{1.5pt}
{\footnotesize
\begin{tabular}{cccccc}
Input & Dense & 10\% & 30\% & 50\% & 70\% \\
\includegraphics[width=.15\linewidth]{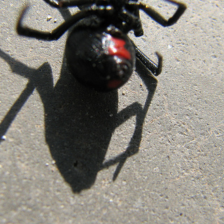} &
\includegraphics[width=.15\linewidth]{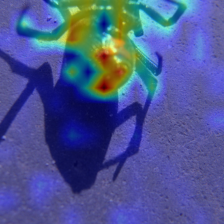} &
\includegraphics[width=.15\linewidth]{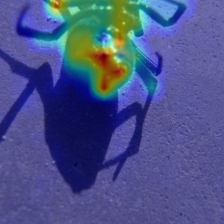} &
\includegraphics[width=.15\linewidth]{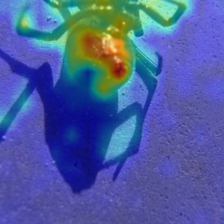} &
\includegraphics[width=.15\linewidth]{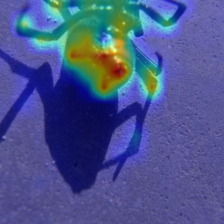} &
\includegraphics[width=.15\linewidth]{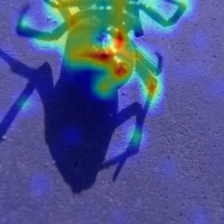} \\
\includegraphics[width=.15\linewidth]{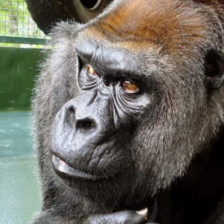} &
\includegraphics[width=.15\linewidth]{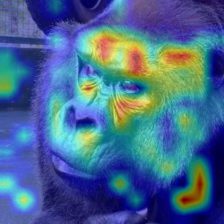} &
\includegraphics[width=.15\linewidth]{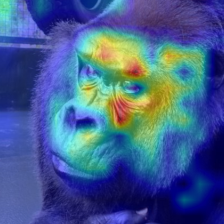} &
\includegraphics[width=.15\linewidth]{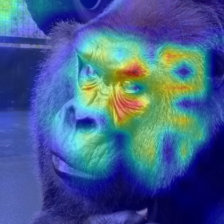} &
\includegraphics[width=.15\linewidth]{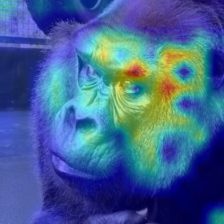} &
\includegraphics[width=.15\linewidth]{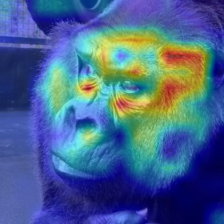} \\
\includegraphics[width=.15\linewidth]{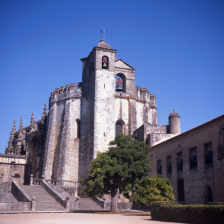} &
\includegraphics[width=.15\linewidth]{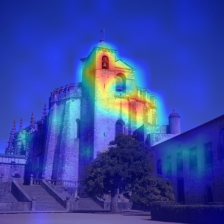} &
\includegraphics[width=.15\linewidth]{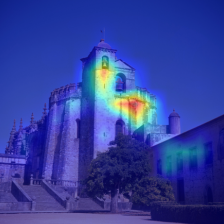} &
\includegraphics[width=.15\linewidth]{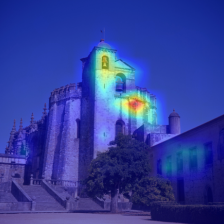} &
\includegraphics[width=.15\linewidth]{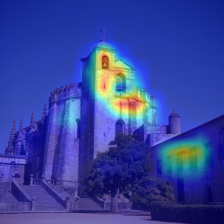} &
\includegraphics[width=.15\linewidth]{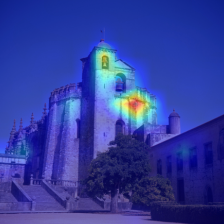} \\
\includegraphics[width=.15\linewidth]{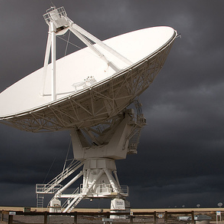} &
\includegraphics[width=.15\linewidth]{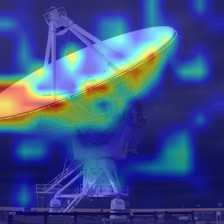} &
\includegraphics[width=.15\linewidth]{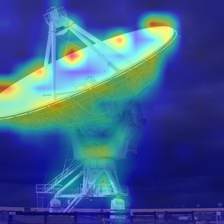} &
\includegraphics[width=.15\linewidth]{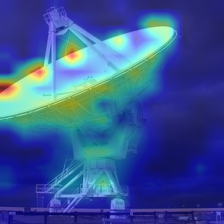} &
\includegraphics[width=.15\linewidth]{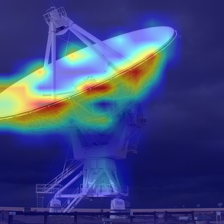} &
\includegraphics[width=.15\linewidth]{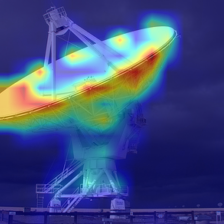} \\
\end{tabular}}
\caption{Target-class attribution maps of the dense model and of Wanda-pruned models with CE recovery at increasing sparsity, for four validation images (black widow, gorilla, monastery, radio telescope); each map is scaled to $[0,1]$ separately.}
\label{fig:qualitative}
\end{figure}

\cref{fig:qualitative} shows target-class transformer-attribution maps of the dense model and of Wanda-pruned models (CE recovery) at increasing sparsity.

\end{document}